\documentclass{article}

\usepackage{microtype}
\usepackage{booktabs} % for professional tables
\usepackage{hyperref}

\usepackage[accepted]{icml2018}
\usepackage{Definitions}
\usepackage{algorithm}
\usepackage{algorithmic}
\usepackage{graphicx}
\usepackage{subfigure}
\usepackage{caption}
\usepackage{natbib}

\usepackage{enumitem}
\renewcommand{\algorithmicrequire}{\textbf{Input:}}
\renewcommand{\algorithmicensure}{\textbf{Output:}}

\usepackage{changes}
\usepackage{color}

\newcommand{\xlm}[1]{{\color{black}{#1}}}

\icmltitlerunning{Towards Black-box Iterative Machine Teaching}

\begin{document}
	\twocolumn[
    \icmltitle{Towards Black-box Iterative Machine Teaching}

	\icmlsetsymbol{equal}{*}

\begin{icmlauthorlist}
\icmlauthor{Weiyang Liu}{equal,gt}
\icmlauthor{Bo Dai}{equal,gt}
\icmlauthor{Xingguo Li}{um}
\icmlauthor{Zhen Liu}{gt}
\icmlauthor{James M. Rehg}{gt}
\icmlauthor{Le Song}{gt,af}
\end{icmlauthorlist}

\icmlaffiliation{gt}{Georgia Tech}
\icmlaffiliation{um}{University of Minnesota}
\icmlaffiliation{af}{Ant Financial}
\icmlcorrespondingauthor{W.~L.}{wyliu@gatech.edu}
%\icmlcorrespondingauthor{Eee Pppp}{ep@eden.co.uk}

\icmlkeywords{Machine Teaching, Blacl-box, Cross-space, Active Teaching, Convergence}

\vskip 0.3in
]

\printAffiliationsAndNotice{\icmlEqualContribution}

\begin{abstract}
		In this paper, we make an important step towards the black-box machine teaching by considering the cross-space machine teaching, where the teacher and the learner use different feature representations and the teacher can not fully observe the learner's model. In such scenario, we study how the teacher is still able to teach the learner to achieve faster convergence rate than the traditional passive learning. We propose an active teacher model that can actively query the learner (\ie, make the learner take exams) for estimating the learner's status and provably guide the learner to achieve faster convergence. The sample complexities for both teaching and query are provided. In the experiments, we compare the proposed active teacher with the omniscient teacher and verify the effectiveness of the active teacher model.
\end{abstract}

\vspace{-3.97mm}
\section{Introduction}
Machine teaching \cite{zhu2015machine,zhu2013machine,zhu2018overview} is the problem of constructing a minimal dataset for a target concept such that a student model (\ie, leaner) can learn the target concept based on this minimal dataset. Recently, machine teaching has been shown very useful in applications ranging from human computer interaction \cite{suh2016label}, crowd sourcing \cite{singla2014near,singla2013actively} to cyber security \cite{alfeld2016data,alfeld2017explicit}. Besides various  applications, machine teaching also has nice connections with curriculum learning \cite{bengio2009curriculum,hinton2015distilling}. In traditional machine learning, a teacher usually constructs a batch set of training samples, and provides them to a student in one shot without further interactions. Then the student keeps learning from this batch dataset and tries to learn the target concept. Previous machine teaching paradigm \cite{zhu2013machine,zhu2015machine,liu2016teaching} usually focuses on constructing the smallest such dataset, and characterizing the size of such dataset, called the \emph{teaching dimension} of the student model.
\par
\begin{figure}[t]
		\centering
		\footnotesize
		\renewcommand{\captionlabelfont}{\footnotesize}
		\includegraphics[width=2.6in]{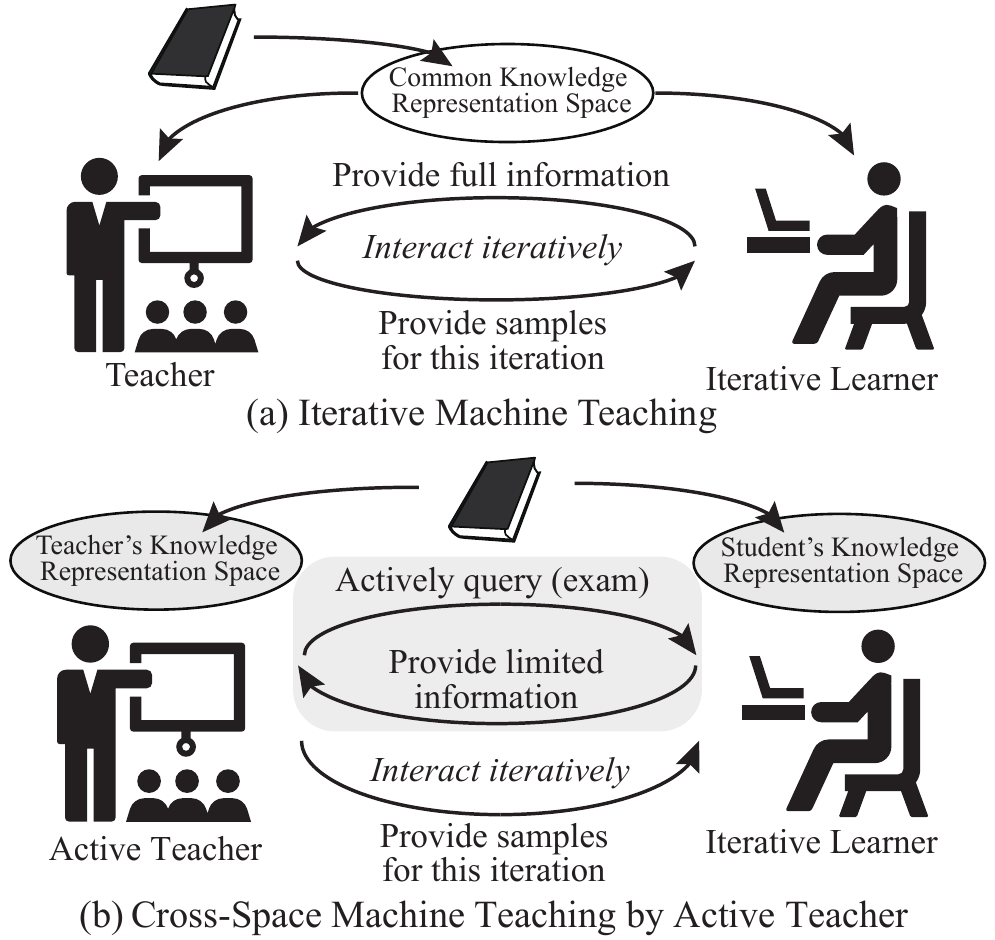}
		\vspace{-3mm}
		\caption{\footnotesize Comparison between iterative machine teaching and cross-space machine teaching by active teacher.}\label{fig1}
		\vspace{-7mm}
\end{figure}
For machine teaching to work effectively in practical scenarios, \cite{liu2017iterative} propose an iterative teaching framework which takes into consideration that the learner usually uses iterative algorithms (e.g. gradient descent) to update the models. Different from the traditional machine teaching framework where the teacher only interacts with the student in one-shot, the iterative machine teaching allows the teacher to interact with the student in every single iteration. It hence shifts the teaching focus from models to algorithms: the objective of teaching is no longer constructing a minimal dataset in one shot but searching for samples so that the student learns the target concept in a minimal number of iterations (\ie, fastest convergence for the student algorithm). Such a minimal number of iterations is called \xlm{the} \emph{iterative teaching dimension} for the student algorithm. \cite{liu2017iterative} mostly consider the simplest iterative case where the teacher can fully observe the student. This case is interesting in theory but too restrictive in practice.
	\par
\vspace{-.8mm}
	Human teaching is arguably the most realistic teaching scenario in which the learner is completely a black-box to the teacher. Analogously, the ultimate problem for machine teaching is how to teach a black-box learner. We call such problem \emph{black-box machine teaching}. Inspired by the fact that the teacher and the student typically represent the same concept but in different ways, we present a step towards the black-box machine teaching -- \emph{cross-space machine teaching}, where the teacher {\bf i)} does not share the same feature representation with the student, and {\bf ii)} can not observe the student model. This setting is interesting in the sense that it can both relax the assumptions for iterative machine teaching and improve our understanding on human learning.
	\par
\vspace{-.8mm}
	Inspired by a real-life fact, that a teacher will regularly examine the student to learn how well the student has mastered the concept, we propose an active teacher model to address the cross-space teaching problem. The active teacher is allowed to actively query the student with a few (limited) samples every certain number of iterations, and the student can only return the corresponding prediction results to the teacher. For example, if the student uses a linear regression model, it will return to the teacher its prediction $\langle w^t,\tilde{x} \rangle$ where $w^t$ is the student parameter at the $t$-th iteration and $\xtil$ is the representation of the query example in student's feature space. Under suitable conditions, we show that the active teacher can always achieve faster rate of improvement than a random teacher that feeds samples randomly. In other words, the student model guided by the active teacher can provably achieve faster convergence than the stochastic gradient descent (SGD). Additionally, we discuss the extension of the active teacher to deal with the learner with forgetting behavior, and the learner guided by multiple teachers.
	\par
\vspace{-.8mm}
	To validate our theoretical findings, we conduct extensive experiments on both synthetic data and real image data. The results show the effectiveness of the active teacher.
	
\vspace{-2mm}
\section{Related Work}
\vspace{-1mm}

	Machine teaching defines a task where we need to find an optimal training set given a learner and a target concept. \cite{zhu2015machine} describes a general teaching framework which has nice connections to curriculum learning \cite{bengio2009curriculum} and knowledge distillation~\cite{hinton2015distilling}. \cite{zhu2013machine} considers Bayesian learners in exponential family and formulates the machine teaching as an optimization problem over teaching examples that balance the future loss of the learner and the effort of the teacher. \cite{liu2016teaching} give the teaching dimension of linear learners. Machine teaching has been found useful in cyber security \cite{mei2015using}, human computer interaction \cite{meek2016analysis}, and human education \cite{khan2011humans}. \cite{johns2015becoming} extend machine teaching to human-in-the-loop settings. \cite{doliwa2014recursive,gao2015teaching,zilles2008teaching,samei2014algebraic,chen18explain} study the machine teaching problem from a theoretical perspective.
	\par
\vspace{-.8mm}
	Previous machine teaching works usually ignore the fact that a student model is typically optimized by an iterative algorithm (\eg, SGD), and in practice we focus more on how fast a student can learn from the teacher. \cite{liu2017iterative} propose the iterative teaching paradigm and an omniscient teaching model where the teacher knows almost everything about the learner and provides training examples based on the learner's status. Our cross-space teaching serves as a stepping stone towards the black-box iterative teaching.
	
	\vspace{0mm}
	\section{Cross-Space Iterative Machine Teaching}
	\vspace{-1mm}
	
	The cross-space iterative teaching paradigm is different from the standard iterative machine teaching in terms of two major aspects: {\bf i)} the teacher does not share the feature representation with the student; {\bf ii)} the teacher cannot observe the student's current model parameter in each iteration. Specifically, we consider the following teaching settings:
	\par
\vspace{-.8mm}
	\textbf{Teacher.} The teacher model observes a sample $\Acal$ (e.g. image, text, etc.) and represents it as a feature vector $\thickmuskip=2mu \medmuskip=2mu x_{\Acal}\in\mathbb{R}^{d}$ and a label $\thickmuskip=2mu \medmuskip=2mu y\in\mathbb{R}$. The teacher knows the model (\eg, loss function) and the optimization algorithm (including the learning rate\footnote{For simplicity, the teacher is assumed to know the learning rate of the learner, but this prior is not necessary, as discussed later.}) of the learner, and the teacher \xlm{preserves} an optimal parameter $v^*$ of this model in its own feature space. We denote the prediction of the teacher as $\thickmuskip=2mu \medmuskip=2mu \hat{y}_{v^\ast} = \langle v^\ast,x\rangle$\footnote{For simplicity, we omit the bias term throughout the paper. It is straightforward to add them back.}.
	\par
\vspace{-.8mm}
	\textbf{Learner.} The learner observes the same sample $\Acal$ and represents it as a vectorized feature $\thickmuskip=2mu \medmuskip=2mu \tilde{x}_{\Acal}\in\mathbb{R}^{s}$ and a label $\thickmuskip=2mu \medmuskip=2mu \tilde{y}\in\mathbb{R}$. The learner uses a linear model $\langle w,\tilde{x} \rangle$ where $w$ is its model parameter and updates it with SGD (if guided by a passive teacher). We denote the prediction of the student model as $\thickmuskip=2mu \medmuskip=2mu \hat{y}_{w}^t = \langle w^t, \tilde{x}\rangle$ in $t$-th iteration.
	\par
\vspace{-.8mm}
	\textbf{Representation.} Although the teacher and learner do not share the feature representation, we still assume their representations have an intrinsic relationship. For simplicity, we assume there exists a unknown one-to-one mapping $\Gcal$ from the teacher's feature space to the student's feature space such that $\thickmuskip=2mu \medmuskip=2mu \tilde{x}=\Gcal(x)$. However, the conclusions in this paper are also applicable to injective mappings.
	Unless specified, we assume that $y=\tilde{y}$ by default.
	\par
\vspace{-.8mm}
	\textbf{Interaction.} In each iteration, the teacher will provide a training example to the learner and the learner will update its model using this example. The teacher cannot directly observe the model parameter $w$ of the student. In this paper, the active teacher is allowed to query the learner with a few examples every certain number of iterations. The learner can only return to the teacher its prediction $\langle w^t, \xtil\rangle$ in the regression scenario, its predicted label $\thickmuskip=2mu \medmuskip=2mu \textnormal{sign}(\langle w^t, \xtil\rangle)$ or confidence score $S(\langle w^t,\xtil\rangle)$ in the classification scenario,
	where $w^t$ is the student's model parameter \xlm{at} $t$-th iteration and $S(\cdot)$ is some nonlinear function. Note that the teacher and student preserve the same loss function $\ell(\cdot,\cdot)$.
	\par
\vspace{-1mm}
	Similar to~\cite{liu2017iterative}, we consider three ways for the teacher to provide examples to the learner:
	\vspace{-0.5mm}
	\par
	\textbf{Synthesis-based teaching.} In this scenario, the space of provided examples is
	\vspace{-1.5mm}
	\begin{equation}
	\footnotesize
	\begin{aligned}
	\Xcal &= \{x\in \RR^d, \nbr{x}\le R\}\\
	\Ycal &= \RR\text{ (Regression) or } \cbr{-1, 1}\text{ (Classification)}. \nonumber
	\end{aligned}
	\end{equation}
	\par
	\vspace{-3mm}
	{\bf Combination-based teaching.} In this scenario, the space of provided examples is ($\alpha_i\in\mathbb{R}$)
	\vspace{-2mm}
	\begin{equation}
	\footnotesize
	\begin{aligned}
	\Xcal &= \big{\{} x | \ \|x\|\leq R, x = \Sigma_{i=1}^k \alpha_i x_i, x_i \in \Dcal \big{\}}, \Dcal = \cbr{x_1, \ldots, x_k} \\
	\Ycal &= \RR\text{ (Regression) or } \cbr{-1, 1}\text{ (Classification)} \nonumber
	\end{aligned}
	\end{equation}
	\par
	\vspace{-3mm}
	{\bf Rescalable pool-based teaching.} This scenario further restrict the knowledge pool for samples. The teacher can pick examples from $\thickmuskip=2mu \medmuskip=2mu \Xcal\times\Ycal$:
	\vspace{-2mm}
	\begin{equation}
	\footnotesize
	\begin{aligned}
	\Xcal & = \{ x | \ \|x\|\leq R,  x = \gamma x_i, x_i \in \Dcal, \gamma\in\RR \}, \Dcal= \{ x_1, \ldots\}\\
	\Ycal &= \RR\text{ (Regression) or } \cbr{-1, 1}\text{ (Classification)} \nonumber
	\end{aligned}
	\end{equation}
	\par
	\vspace{-3mm}
	We also note that the pool-based teaching (without rescalability) is the most restricted teaching scenario and it is very close to the practical settings.

	\vspace{-3mm}
	\section{The Active Teaching Algorithm}
	\vspace{-1mm}
	
	To address the cross-space iterative machine teaching, we propose the active teaching algorithm, which actively queries its student for its prediction output. We first describe the general version of the active teaching algorithm. Then without loss of generality, we will discuss three specific examples: least square regression (LSR) learner for regression, logistic regression (LR) and support vector machine~(SVM) learner for classification \xlm{\cite{friedman2001elements}.}

	\vspace{-3mm}
	\subsection{General Algorithm}
	\vspace{-2mm}
	\begin{figure}[ht]
		\centering
		\renewcommand{\captionlabelfont}{\footnotesize}
		\footnotesize
		\includegraphics[width=2.75in]{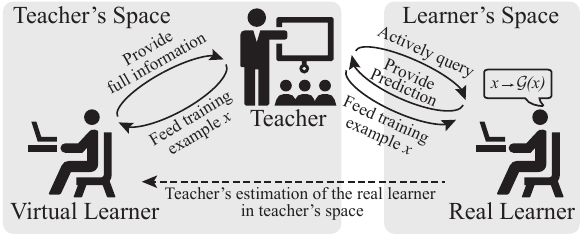}
		\vspace{-3mm}
		\caption{\footnotesize The cross-space teaching by active teacher. The real learner receives training example $x$ but will perceive it as $\Gcal(x)$.}\label{fig2}
		\vspace{-2.5mm}
	\end{figure}

	Inspired by human teaching, we expand the teacher's capabilities by enabling the teacher to actively query the student. The student will return its predictions to the teacher. Based on the student's feedback, The teacher will estimate the student's status and determine which example to provide next time. The student's feedback enables the active teacher to teach without directly observing the student's model.

	The active teacher can choose to query the learner with a few samples in each iteration, and the learner will usually report the prediction $F(\langle w,\tilde{x} \rangle)$ where $F(\cdot)$ denotes some function of the inner product prediction. For example, we usually have $F(z)=z$ for regression and $\thickmuskip=2mu \medmuskip=2mu F(z)=\textnormal{sign}(z)$ or $\thickmuskip=2mu \medmuskip=2mu F(z)=\frac{1}{1 + \exp(-z)}$ for classification. Based on our assumption that there is an unknown mapping from teacher's feature to student's feature, there also exists a mapping from the model parameters of the teacher to those of the student. These active queries enables the teacher to estimate the student's corresponding model parameter ``in the teacher's space" and maintain a \emph{virtual learner}, the teacher's estimation of the real learner, in its own space. The teacher will decide which example to provide based on its current virtual learner model. The ideal virtual learner $v$ will have the same prediction output as the real learner, \ie\ $\thickmuskip=2mu \medmuskip=2mu \inner{v}{x}=\inner{w}{\tilde{x}}$ where $\thickmuskip=2mu \medmuskip=2mu \tilde{x}=\Gcal(x)$. Equivalently, $\thickmuskip=2mu \medmuskip=2mu v=\Gcal^\top(w)$ always holds for the ideal virtual learner, where $\Gcal^\top$ is the conjugate mapping of $\Gcal$. Note that for the purpose of analysis, we assume that $\Gcal$ is a generic linear operator, though our analysis can easily extends to general cases. In fact, one of the most important challenges in active teaching is to recover a virtual student that approximates the real leaner as accurately as possible.
	The estimation error of \xlm{the} teacher may affect the quality of training examples that the teacher provides for the real learner. Intuitively, if we can recover the virtual learner with an appropriate accuracy, then we can still achieve faster teaching speed than that of passive learning.
	Fig. \ref{fig2} shows the pipeline of the cross-space teaching.
    \vspace{-1mm}
	\par
	With full access to the obtained virtual learner in the teacher's space, the teacher can perform omniscient teaching as in \cite{liu2017iterative}. Specifically, the active teacher will optimize the following objective:
	\vspace{-1mm}
	\begin{equation}\label{T1T2_min}
	\scriptsize
	\begin{aligned}
	\hspace{-0.1in}\argmin_{x\in\Xcal,y\in\Ycal} \eta_t^2 \nbr{\frac{\partial \ell(\inner{v^t}{x},y)}{\partial v^t}}_2^2- 2\eta_t \inner{v^t - v^\ast}{\frac{\partial \ell(\inner{v^t}{x},y)}{\partial v^t}}\hspace{-0.1in}
	\end{aligned}
	\vspace{-1mm}
	\end{equation}
	where \xlm{$\ell$ is a loss function and} $v^t$ is the teacher's estimation of ${\Gcal}^\top(w^t)$ after the teacher \xlm{performs an} active query in $t$-th iteration (\ie, the current model parameter of the virtual learner).
	$\eta_t$ is the learning rate of the virtual learner.
	The learning rate of the student model is not necessarily needed. The general teaching algorithm is given in Algorithm~\ref{alg1}.
\par
\vspace{-0.5mm}
	Particularly, different types of feedback (\ie, the form of $F(\cdot)$) from learners contain different amount of information, resulting in different levels of difficulties in recovering the parameters of the learner's model. We will discuss two general ways to recover the virtual learner for two types of frequently used feedbacks in practice.
	
	{
		\begin{algorithm}[!t]
			\small
			\caption{The active teacher}
			\begin{algorithmic}[1]\label{alg1}
				\STATE Randomly initialize the student parameter $w^0$;
				\STATE Set $t = 1$, $\textnormal{exam}=\textnormal{True}$ (\ie, whether we make the student takes exams) and maximal iteration number $T$;
				\WHILE {$v^t$ has not converged or $t<T$}
				\IF{$\Gcal^\top\Gcal\neq I$ and $\textnormal{exam}=\textnormal{True}$}
				\STATE Obtain an estimation $\hat{\Gcal}^\top(w^t)$ of the student model in the teacher's space using the virtual learner construction Algorithm~\ref{alg2};
				\STATE $v^t = \hat{\Gcal}^\top(w^t)$;
				\ELSIF{$\Gcal^\top\Gcal= I$ and $\textnormal{exam}=\textnormal{True}$}
				\STATE Perform the one-time ``background'' exam using Algorithm~\ref{alg2} and set $\textnormal{exam}$ to $\textnormal{False}$;
				\ENDIF
				\STATE Solve the optimization for the virtual learner (\eg\ pool-based teaching):
				\begin{equation*}
				\setlength{\abovedisplayskip}{1mm}
				\setlength{\belowdisplayskip}{1mm}
				\footnotesize
				\begin{aligned}
				(x^t,y^t) = &\argmin_{x\in\mathcal{X},y\in\mathcal{Y}}~\eta_t^2 \nbr{  \frac{\partial \ell\rbr{\inner{v^{t-1}}{x}, y}}{\partial v^{t-1}}}^2 \- \\[-1mm]
				&\hspace{0.1in}-2\eta_t \inner{v^{t-1} - v^\ast}{ \frac{\partial \ell\rbr{\inner{v^{t-1}}{ x},y}}{\partial v^{t-1}}}
				\end{aligned}
				\end{equation*}
				\IF{$\textnormal{exam}=\textnormal{False}$}
					\STATE Use the selected example $(x^t,y^t)$ to perform the update of the virtual learner in the teacher's space:
					\begin{equation*}
					\setlength{\abovedisplayskip}{1mm}
					\setlength{\belowdisplayskip}{1mm}
					\footnotesize
					\begin{aligned}
					v^{t} = v^{t-1} - \eta_t\, \frac{\partial \ell\rbr{\inner{v^{t-1}}{x^t},y^t}}{\partial v^{t-1}}.
					\end{aligned}
					\end{equation*}
				\ENDIF
				\STATE Use the selected example $(\tilde{x}^t,\tilde{y}^t)$ where $\thickmuskip=2mu \medmuskip=2mu \tilde{x}=\Gcal(x),\tilde{y}=y$ to perform the update of the real learner in the student's space:
				\begin{equation*}
				\setlength{\abovedisplayskip}{1mm}
				\setlength{\belowdisplayskip}{1mm}
				\footnotesize
				\begin{aligned}
				w^{t} = w^{t-1} - \eta_t\, \frac{\partial \ell\rbr{\inner{w^{t-1}}{\tilde{x}^t},\tilde{y}^t}}{\partial w^{t-1}}.
				\end{aligned}
				\end{equation*}
				\STATE $t\leftarrow t+1$;
				\ENDWHILE
			\end{algorithmic}
		\end{algorithm}
	}
	\par
\vspace{-1.1mm}
	\textbf{Exact recovery of the virtual learner.} We know that the learner returns a prediction in the form of $F(\inner{w}{\tilde{x}})$. In general, if $F(\cdot)$ is an one-to-one mapping, we can exactly recover the ideal virtual learner (\ie\ $\Gcal^\top(w)$) in the teacher's space using the system of linear equations. In other words, the recovery of virtual learner could be exact as long as there is no information loss from $\inner{w}{\tilde{x}}$ to $F(\inner{w}{\tilde{x}})$. Specifically, we have $\thickmuskip=2mu \medmuskip=2mu \inner{v}{q_j}=\inner{w}{\tilde{q_j}}$ where $q_j$ is the $j$-th query for the learner. Because $\inner{w}{\tilde{q_j}}$ is given by the real learner, we only need to construct $d$ queries ($d$ is the dimension of the teacher space) and require $\{q_1,q_2,\cdots,q_d\}$ to be linearly independent to estimate $v$. Without no numerical error, we can exactly recover $v$. Since the recovery is exact, we have $\thickmuskip=2mu \medmuskip=2mu \Gcal^\top(w)=v$. Note that there are cases that we can achieve exact recovery without $F(\cdot)$ being an one-to-one mapping. For example, $F(z)=\max(0,z)$ (hinge function) is not an one-to-one mapping but we can still achieve exact recovery.
	\par
	\vspace{-1.1mm}
	\textbf{Approximate recovery of the virtual learner.} If $F(\cdot)$ is not an one-to-one mapping (\eg, $\textnormal{sign}(\cdot)$ which provides $1$-bit feedback), then generally we may not \xlm{be} able to exactly recover the student's parameters. Therefore, we have to develop a more intelligent technique (\ie\ less sample complexity) to estimate $\Gcal^\top(w)$. In this paper, we use active learning \cite{settles2010active} to help the teacher better estimate $\Gcal^\top(w)$ for the virtual learner. One of the difficulties is that the active learning algorithm obtains the parameters of a model based on the predicted labels on which the norm of the weights has no effect. It becomes ambiguous which set of weights the teacher should choose. Therefore, the active teacher also needs to have access to the norm of the student's weights for recovering the virtual learner. In the following sections, we will develop and analyze our estimation algorithm for the virtual learner based on the existing active learning algorithms with guarantees on sample complexity  \cite{balcan2009agnostic,ailon2012active,hanneke2007bound,schein2007active,settles2010active}.
    \par
    {
    	\begin{algorithm}[!t]
    		\small
    		\caption{The virtual learner construction}
    		\begin{algorithmic}[1]\label{alg2}
    			\IF{The feedback function $F(z)$ is an one-to-one mapping or a hinge function}
    			\STATE Perform one-time exam by actively query multiple examples;
    			\STATE Solve a system of linear equations to obtain the exact recovery of the ideal virtual learner;
    			\ELSE
    			\STATE Apply acitve learning algorithms to perform an approximate recovery of the ideal virtual learner (in this case, the teacher will need to know the norm of the student model);
    			\ENDIF
    		\end{algorithmic}
    	\end{algorithm}
    }

	\vspace{-2mm}
	\subsection{Least Square Regression Learner}
	\vspace{-1mm}

	For the LSR learner, we use the following model:
\vspace{-2mm}
	\begin{equation}
	\footnotesize
	\begin{aligned}
	\min_{w\in\mathbb{R}^s,b\in\mathbb{R}}\frac{1}{n}\sum_{i=1}^{n}\frac{1}{2}(\langle w,\tilde{x}_i\rangle-\tilde{y}_i)^2.
	\end{aligned}
	\end{equation}
	Because $\thickmuskip=2mu \medmuskip=2mu F(\inner{w}{\tilde{x}})=\inner{w}{\tilde{x}}$, the LSR learner belongs to the case where the active teacher can exactly recover the ideal virtual learner. When $\Gcal^\top\Gcal = I$, the teacher only need to perform active exam once.
	It can be viewed as a ``background exam'' for the teacher to figure out how well the student has mastered the knowledge at the beginning, and the teacher can track the dynamics of students exactly later. Otherwise, for \xlm{a} general one-to-one mapping $\Gcal$, the teacher needs to query the student in each iteration. Still, the teacher can reuse the same set of queries in all iterations.
	
	\vspace{-2mm}
	\subsection{Logistic Regression Learner}
	\vspace{-1mm}
	
	For the LR learner, we use the following model (without loss of generality, we consider the binary classification):
	\begin{equation}
		\setlength{\abovedisplayskip}{1mm}
		\setlength{\belowdisplayskip}{1mm}
	\footnotesize
	\begin{aligned}
	\min_{w\in\mathbb{R}^s,b\in\mathbb{R}}\frac{1}{n}\sum_{i=1}^{n}\log\big(1+\exp\{-\tilde{y}_i(\langle w,\tilde{x}_i\rangle)\}\big)
	\end{aligned}
	\end{equation}
	We discuss two cases separately: (1) the learner returns the probability of each class (\ie\ $F(z)=S(z)$ where $S(\cdot)$ denotes a sigmoid function); (2) the learner only returns the predicted label (\ie\ $F(z)=\textnormal{sign}(z)$).
	\par
\vspace{-0.5mm}
	In the first case where $F(\cdot)$ is a sigmoid function, we can exactly recover the ideal virtual learner. This case is essentially similar to the LSR learner where we need only one ``background exam'' if $\thickmuskip=2mu \medmuskip=2mu \Gcal^\top\Gcal = I$ and we can reuse the queries in each iteration for \xlm{a} general one-to-one mapping $\Gcal$ ($\thickmuskip=2mu \medmuskip=2mu \Gcal^\top\Gcal \neq I$). In the second case where $F(\cdot)$ is a sign function, we can only approximate the ideal virtual learner with some error. In this case, we use active learning to do the recovery.
	
	\vspace{-2mm}
	\subsection{Support Vector Machine Learner}
	\vspace{-1mm}
	
	For the SVM learner, we use the following model for the binary classification:
	\begin{equation}
	\setlength{\abovedisplayskip}{1mm}
	\setlength{\belowdisplayskip}{1mm}
	\footnotesize
	\begin{aligned}
	\min_{w\in\mathbb{R}^s,b\in\mathbb{R}}\frac{1}{n}\sum_{i=1}^{n}\max(1-y_i(w^T\tilde{x}_i+b),0)
	\end{aligned}
	\end{equation}
	Similarly, we have two cases: (1) the learner returns the hinge value of each class (\ie\ $\thickmuskip=2mu \medmuskip=2mu F(z)=\max(0,z)$; (2) the learner only returns the label (\ie\ $F(z)=\textnormal{sign}(z)$).
	\par
\vspace{-0.5mm}
	In the first case where $F(\cdot)$ is a hinge function, we can still recover the ideal virtual learner. Although the hinge function is not a bijective mapping (only half of it is one-to-one), we \xlm{prove that it can still} achieve exact recovery with slightly more query samples. \xlm{For $\Gcal^\top\Gcal=I$, we need only one ``background exam''} as in the case of the LR learner. Otherwise, we still need to query the student in each iteration. In the second case where $F(\cdot)$ is a sign function, we can only approximate the ideal virtual learner with some error.

	\vspace{-3mm}
	\section{Theoretical Results}
	\vspace{-1mm}
	
	We define an important notion of being ``exponentially teachable'' to characterize the teacher's performance.
	\vspace{-2mm}
	\begin{definition}\label{def:et}
		Given $\epsilon>0$, the loss function $\ell$ and feature mapping $\Gcal$, $(\ell, \Gcal)$ is \textbf{exponentially teachable} (ET) if the number of total samples (teaching samples and query samples) is $t = \Ocal({\rm poly}(\log \frac{1}{\epsilon}))$ for a learner to achieve $\epsilon$-approximation, i.e., $\nbr{\Gcal^\top (w^t) - v^\ast} \leq \epsilon$.
	\end{definition}
\vspace{-2mm}
	Note that the potential dependence of $t$ on the problem dimension is omitted here, which will be discussed in detail in the following.
	We summarize our theoretical results in Table \ref{thm_res}. Given a learner that is exponentially teachable by the omniscient teacher, we find that the learner \xlm{is not} exponentially teachable by the active teacher only when $F(\cdot)$ is not an one-to-one mapping and the teacher uses rescalable pool-based teaching.
	\begin{table}[h]
		\centering
		\vspace{-2mm}
		\scriptsize
		\renewcommand{\captionlabelfont}{\footnotesize}
		\newcommand{\tabincell}[2]{\begin{tabular}{@{}#1@{}}#2\end{tabular}}
		\setlength{\abovecaptionskip}{4pt}
		\setlength{\belowcaptionskip}{-5pt}
		\begin{tabular}{|c|c|c|c|}
			\hline
			$F(\cdot)$ & \tabincell{c}{Synthesis\\teaching} & \tabincell{c}{Combination\\teaching} & \tabincell{c}{Rescalable\\pool teaching} \\
			\hline
			\tabincell{c}{One-to-one or hinge function} & $\surd$ & $\surd$ & $\surd$  \\
			\hline
			\tabincell{c}{The other function} & $\surd$ & $\surd$ & $\times$  \\
			\hline
		\end{tabular}
		\caption{\footnotesize The exponential teachability by  active teacher. Assume that the learner is exponentially teachable by omniscient teacher.}\label{thm_res}
	\end{table}

    \vspace{-5mm}
    \subsection{Synthesis-Based Active Teaching}
    \vspace{-1.5mm}

    We denote
    $ \sigma_{\max} = \max_{x^\top x=1} \Gcal^\top(x) \Gcal(x)$ and $ \sigma_{\min} = \min_{x^\top x=1} \Gcal^\top(x) \Gcal(x) >0 $ ($\Gcal$ is invertible).
    We first discuss the teaching algorithm when the teacher is able to exactly recover the student's parameters. A generic theory for synthesis-based ET is provided as follows.
    \vspace{-1mm}
	\begin{theorem}\label{thm:exact_recovery}
		\xlm{Suppose that the teacher can recover $\Gcal^\top (w^t)$ exactly using $m$ samples at each iteration. If for any $\thickmuskip=2mu \medmuskip=2mu v\in\RR^d$, there exists $\thickmuskip=2mu \medmuskip=2mu \gamma\neq 0$ and $\hat y$ such that $\thickmuskip=2mu \medmuskip=2mu \hat x = \gamma\rbr{v - v^*}$ and}
		\begin{equation*}
		\setlength{\abovedisplayskip}{1mm}
		\setlength{\belowdisplayskip}{1mm}
		\footnotesize
		\begin{aligned}
		0 < \gamma\nabla_{\inner{v^t}{\hat x}}\ell\rbr{\inner{v^t}{\hat x}, \hat y} < \frac{2\sigma_{\min}}{\eta\sigma^2_{\max}},
		\end{aligned}
		\end{equation*}
		then $(\ell,\Gcal)$ is ET with $\Ocal\rbr{\rbr{m+1}\log\frac{1}{\epsilon}}$ samples.
	\end{theorem}
	\vspace{-1mm}
	\noindent{\bf Existence of the exponentially teachable $\rbr{\ell, \Gcal}$ via exact recovery.} Different from~\cite{liu2017iterative} where the condition for synthesis-based exponentially teaching is only related to the loss function $\ell$, the condition for the cross-space teaching setting is related to both loss function $\ell$ and feature mapping $\Gcal$. The spectral property of $\Gcal$ is involved due to the differences of feature spaces, leading to the mismatch of parameters of the teacher and student.
	It is easy to see that $\exists~\Gcal$ such that the commonly used loss functions, \eg, absolute loss, square loss, hinge loss, and logistic loss, are ET with exact recovery, i.e., $\Gcal^\top (w^t) = v^t$. This can be shown by construction. For example, if the $\frac{\sigma_{\min}}{\sigma^2_{\max}} = \frac{1}{2}$, the ET condition will be the same for both omniscient teacher~\cite{liu2017iterative} and active teacher.
\par
\vspace{-1mm}
	Next we present generic results of the sample complexity $m$ required to recover $\Gcal^\top$, which is a constant to $\epsilon$ (i.e., $(\ell,\Gcal)$ is ET), as follows.
	\vspace{-2mm}
	\begin{lemma}\label{thm:invertible_F}
		If \xlm{$F(\cdot)$ is} bijective, then we can exactly recover $\thickmuskip=2mu \medmuskip=2mu \Gcal^\top (w)\in\RR^d$ with $d$ samples.
	\end{lemma}
	\vspace{-4mm}
	\begin{lemma}\label{thm:hinge_F}
		If $F(\cdot) = \max\rbr{0, \cdot}$, then we can exactly recover $\Gcal^\top (w)\in \RR^d$ with $2d$ samples.
	\end{lemma}
	\vspace{-2mm}
	Lemma~\ref{thm:invertible_F} and \ref{thm:hinge_F} cover $\thickmuskip=2mu \medmuskip=2mu F(\cdot) = I(\cdot)$, $\thickmuskip=2mu \medmuskip=2mu F(\cdot) = S(\cdot)$, or $\thickmuskip=2mu \medmuskip=2mu F(\cdot) = \max\rbr{0, \cdot}$, where $I$ denotes the identity mapping and $S$ denotes some sigmoid function, \eg, logistic function, hyperbolic tangent, error function\xlm{, etc.}
	If the student's answers to the queries \xlm{via} these student feedbacks $F(\cdot)$ in the exam phase, then we can exactly recover $\thickmuskip=2mu \medmuskip=2mu v = \Gcal^\top (w)\in\RR^d$ with arbitrary $d$ independent data, omitting the numerical error. Also note that the query samples in Lemma~\ref{thm:invertible_F} and \ref{thm:hinge_F} can be reused in each iteration, thus the query sample complexity is $m=\Ocal(d)$, which is formalized as follows.
	\vspace{-2mm}
	\begin{corollary}\label{cor:identity_sigmoid_F}
		Suppose that the student answers questions in query phase via $F(\cdot) = I(\cdot)$, $F(\cdot) = S(\cdot)$, or $F(\cdot) = \max\rbr{0, \cdot}$, then $(\ell,\Gcal)$ is ET with $\Ocal\rbr{\log\frac{1}{\epsilon}}$ teaching samples and $\Ocal(d)$ query samples via exact recovery.
	\end{corollary}
	\vspace{-2mm}
	Here we emphasize that the number of query samples (\ie\ active queries) does not depend on specific tasks. For both regression and classification, as long as the student feedbacks $F(\cdot)$ are bijective functions, \xlm{then Corollary}~\ref{cor:identity_sigmoid_F} holds. The loss function only affects the synthesis or selection of the teaching samples.
\par
\vspace{-1mm}
	In both regression and classification, if $F(\cdot) = \sgn(\cdot)$ which only provides $1$-bit feedback, $F^{-1}$ no longer exists and the exact recovery of $\Gcal^\top(w)$ may not be obtained. In such case, the teacher may only approximate the student's parameter using active learning. We first present the generic result for ET via approximate recovery as follows.
	\vspace{-2mm}
	\begin{theorem}\label{theorem:recursion_with_error}
		Suppose that the loss function $\ell$ is $L$-Lipschitz smooth in a compact domain $\thickmuskip=2mu \medmuskip=2mu \Omega_v\subset\RR^d$ containing $v^\ast$ and sample candidates $(x, y)$ are from bounded set $\thickmuskip=2mu \medmuskip=2mu \Xcal\times\Ycal$, where $\thickmuskip=2mu \medmuskip=2mu \Xcal = \cbr{x\in \RR^d, \nbr{x}\le R}$. Further suppose at $t$-th iteration, the teacher estimates the student $\thickmuskip=2mu \medmuskip=2mu \epsilon_{\text{est}} := \nbr{\Gcal^\top (w^t) - v^t} = \Ocal\rbr{\epsilon}$ with probability at least $1-\delta$ using $m\rbr{\epsilon_{\text{est}}, \delta}$ samples. If for any $v\in\Omega_v$, there exists $\thickmuskip=2mu \medmuskip=2mu \gamma\neq 0$ and $\hat y$ such that for $\thickmuskip=2mu \medmuskip=2mu \hat x = \gamma\rbr{v - v^\ast}$, we have
		\begin{equation*}
		\setlength{\abovedisplayskip}{0mm}
		\setlength{\belowdisplayskip}{0mm}
		\footnotesize
		\begin{aligned}
		&0 < \gamma\nabla_{\inner{v^t}{\hat x}}\ell\rbr{\inner{v^t}{\hat x}, \hat y} < \frac{2\rbr{1 - \lambda}\sigma_{\min}}{\eta\sigma^2_{\max}}, \quad \\[-1.8mm]
		&\text{with  }\, 0<\lambda< \min\big(\frac{\kappa\rbr{\Gcal^\top\Gcal}}{\sqrt{2}},\, 1\big),
		\end{aligned}
		\end{equation*}
		then the student can achieve $\epsilon$-approximation of $v^\ast$ with $\Ocal\rbr{\log\frac{1}{\epsilon}\rbr{1 + m\rbr{\lambda{\epsilon}, \frac{\delta}{\log\frac{1}{\epsilon}}}}}$ samples with probability at least $1-\delta$. If $m\rbr{\epsilon_{\text{est}}, \delta} = \Ocal(\log \frac{1}{\epsilon})$, then $(\ell,\Gcal)$ is ET.
	\end{theorem}
	\vspace{-1mm}
	\noindent{\bf Existence of exponentially teachable $\rbr{\ell, \Gcal}$ via approximate recovery.}
	$m\rbr{\epsilon_{\text{est}}, \delta}$ is the number of samples needed for approximately recovering $\Gcal^{\top}(w^t)$ in each iteration. Different from the exact recovery setting where $m$ only depends on the feature dimension, $m\rbr{\epsilon_{\text{est}}, \delta}$ here also depends on how accurately the teacher wants to recover $\Gcal^{\top}(w^t)$ in each iteration ($\epsilon_{\text{est}}$ denotes the estimation error of $\Gcal^{\top}(w^t)$). The condition for exponentially teachable with approximate recovery is related to both $(\ell,\Gcal)$ and the approximation level of the student parameters, \ie, the effect of $\lambda$.
	For example, if the $\frac{\sigma_{\min}}{\sigma^2_{\max}} = 1$ and $\lambda = \frac{1}{2}$, the exponentially teachable condition will be the same for both the omniscient teaching~\cite{liu2017iterative} and active teaching with exact recovery.
\par
\vspace{-1mm}
	For $\thickmuskip=2mu \medmuskip=2mu F(\cdot) = \sgn(\cdot)$, if the student \xlm{provides} $\sgn\rbr{\inner{w}{\Gcal(x)}}$ for the query $x$, it is unlikely to recover $\Gcal^\top (w)$ unless we know ${\nbr{\Gcal^\top(w)}}$. This leads to the following assumption.
	\vspace{-2mm}
	\begin{assumption}\label{eq:1bit_assumption}
		\xlm{The feedback is $1$-bit, \ie\ $\thickmuskip=2mu \medmuskip=2mu F(\cdot)=\sgn(\cdot)$, and the norm of $\Gcal^\top (w)$ is known to teacher. }
	\end{assumption}
	\vspace{-2mm}
	Assumption~\ref{eq:1bit_assumption} is necessary because $\sgn(\cdot)$ is \xlm{scale} invariant. We can\xlm{not} distinguish \xlm{between} $\Gcal^\top (w)$ and $k \cdot \Gcal^\top (w)$ \xlm{for any} $k \in \RR^+$ only with their signs. The following theorem provides the query sample complexity in this scenario.
	\vspace{-2mm}
	\begin{theorem}\label{thm:sign_F}
		\xlm{Suppose that Assumption~\ref{eq:1bit_assumption} holds. Then} with probability at least $\thickmuskip=2mu \medmuskip=2mu 1 - \delta$, then we can recover $\Gcal^\top(w) \in \RR^d$ with $\tilde\Ocal\rbr{\rbr{d^2 + d\log\frac{1}{\delta}}\log\frac{1}{\epsilon}}$ query samples.
	\end{theorem}
	\vspace{-2mm}
	Combining Theorem~\ref{theorem:recursion_with_error} with Theorem~\ref{thm:sign_F}, we have the results for the 1-bit feedback case.
	\vspace{-2mm}
	\begin{corollary}\label{cor:sign_F}
		\xlm{Suppose Assumption~\ref{eq:1bit_assumption} holds. Then} then $(\ell,\Gcal)$ is ET with $\Ocal\rbr{\log\frac{1}{\epsilon}}$ teaching samples and $\tilde\Ocal\rbr{\log\frac{1}{\epsilon}\log\frac{1}{\lambda\epsilon}\rbr{d^2 + d\log\frac{\log \frac{1}{\epsilon}}{\delta}}}$ query samples.
	\end{corollary}
	\vspace{-2mm}
	\noindent{\bf Trade-off between teaching samples and query samples.} There is a delicate trade-off between query sample complexity (in the exam phase) and teaching sample complexity. Specifically, with $\thickmuskip=2mu \medmuskip=2mu \epsilon_{\text{est}} = \Ocal\rbr{\frac{1}{t^2}}$ and $m\rbr{\Ocal\rbr{\frac{1}{t^2}}}$ query samples, we can already achieve the conclusion that $\thickmuskip=2mu \medmuskip=2mu \nbr{\Gcal^\top (w^{t+1}) - v^\ast}^2$ converges in rate $\Ocal\rbr{\frac{1}{t}}$, which makes the number of teaching samples to be $\Ocal\rbr{\frac{1}{\epsilon^2}}$. We emphasize that this rate is the same with the convergence of SGD minimizing strongly convex functions. Note that the teaching algorithm can achieve at least this rate for general convex loss.
	Compared to the number of teaching samples in Corollary~\ref{cor:sign_F}, although the query samples is less, this setting requires much more effort in teaching. Such phenomenon is reasonable in practice in the sense that if the examination is not accurate, the teacher provides the student less effective samples and hence has to teach for more iterations when the teacher cannot accurately evaluate student's performance.

	We remark that if $\Gcal$ is a unitary operator, \ie, $\Gcal^\top\Gcal = I$, we can show that the teacher need only \textit{one} exam. The key insight is that after the first ``background exam'', the teacher can replace the following exams by updating the virtual learner via the same dynamic of the real learner. This is formalized as follows.
	\vspace{-2mm}
	\begin{lemma}\label{lemma:error_preservation}
		\xlm{Suppose} that $\Gcal$ is a unitary operator\xlm{. If} $\nbr{\Gcal^\top (w^0)  - v^0}\le \epsilon$, then $\nbr{\Gcal^\top (w^{t+1}) - v^{t+1}} \le \epsilon$.
	\end{lemma}
	\vspace{-1.5mm}
	Therefore, with \xlm{a} unitary feature mapping, we only need one exam in the whole teaching procedure. It follows that the query sample complexity in theorem~\ref{theorem:recursion_with_error} will be reduced to $\tilde\Ocal\rbr{\log\frac{1}{\lambda\epsilon}\rbr{d^2 + d\log\frac{\log \frac{1}{\epsilon}}{\delta}}}$ via approximate recovery.
	
	\vspace{-2mm}
	\subsection{Combination-Based Active Teaching}
	\vspace{-1mm}
	
	We discuss how the results for synthesis-based active teaching can be extended to the combination-based active teaching. In this scenario, we assume both training and query samples are constructed by linear combination of $k$ samples \xlm{in} $\thickmuskip=2mu \medmuskip=2mu  \Dcal = \cbr{x_i}_{i=1}^k$. We have the following corollaries for both exact recovery and approximate recovery in the sense of
\begin{equation*}
\setlength{\abovedisplayskip}{1mm}
\setlength{\belowdisplayskip}{0.5mm}
	\begin{aligned}
	\xlm{\inner{v_1}{v_2}_{\Dcal}} &:= \sqrt{v_1^\top \Dcal\rbr{\Dcal^\top\Dcal}^{+}\Dcal^\top v_2},~~\text{and} \nonumber \\
	\nbr{v}_{\Dcal}&:=\inner{v}{v}_{\Dcal}.\label{eqn:metric_D}
	\end{aligned}
\end{equation*}
	Note that with the introduced metric, for $v\in \RR^d$, we only consider its component in ${\rm span}\rbr{\Dcal}$ and the components in the null space will be ignored. Therefore, $\forall~ v_1, v_2\in {\rm span}(\Dcal)$ such that $\nbr{v_1}_{\Dcal} = \nbr{v_2}_{\Dcal}$, we have $\thickmuskip=2mu \medmuskip=2mu  v_1^\top x = v_2^\top x = \xlm{\inner{v_1}{x}_{\Dcal}}$ for all $x\in\RR^d$. Then we have the result via exact recovery as follows.
	\vspace{-2mm}
	\begin{corollary}\label{cor:combination_exact}
		Suppose the learner gives feedbacks in query phase by $\thickmuskip=2mu \medmuskip=2mu  F(\cdot) = I(\cdot)$ or $\thickmuskip=2mu \medmuskip=2mu  F(\cdot) = S(\cdot)$, and $\thickmuskip=2mu \medmuskip=2mu  \Gcal^\top (w^0), v^\ast\in {\rm span}\rbr{\Dcal}$. Then $(\ell,\Gcal)$ is ET with $\Ocal\rbr{\log\frac{1}{\epsilon}}$ teaching samples and ${\rm rank}(\Dcal)$ query samples for exact recovery.
	\end{corollary}
	\vspace{-2mm}
	The result via approximate recovery holds analogously to synthesis-based active teaching, given as follows.
	\vspace{-2mm}
	\begin{corollary}\label{cor:combination_approximate}
		Suppose Assumption~\ref{eq:1bit_assumption} holds, the student answers questions in query phase via $\thickmuskip=2mu \medmuskip=2mu F(\cdot) = I(\cdot)$ or $\thickmuskip=2mu \medmuskip=2mu F(\cdot) = S(\cdot)$ and $\thickmuskip=2mu \medmuskip=2mu \Gcal^\top (w^0), v^\ast\in {\rm span}\rbr{\Dcal}$. Then $(\ell,\Gcal)$ is ET with $\Ocal\rbr{\log\frac{1}{\epsilon}}$ teaching samples and $\tilde\Ocal\rbr{\log\frac{1}{\epsilon}\log\frac{1}{\lambda\epsilon}\rbr{d^2 + d\log\frac{\log \frac{1}{\epsilon}}{\delta}}}$ query samples via approximate recovery.
	\end{corollary}
	\vspace{-2mm}
	
	\vspace{-1.5mm}
	\subsection{Rescaled Pool-Based Active Teaching}
	\vspace{-1mm}
	In this scenario, the teacher can only pick examples from a fixed sample candidate pool, $\thickmuskip=2mu \medmuskip=2mu \Dcal = \cbr{x_i}_{i=1}^k$, for teaching and active query. We still evaluate with the metric $\nbr{\cdot}_{\Dcal}$ defined in \eqref{eqn:metric_D}. We first define \emph{pool volume} to characterize the richness of the pool~\cite{liu2017iterative}.
	\vspace{-2mm}
	\begin{definition}[Pool Volume]
		Given the training example pool $\Xcal\in \RR^d$, the volume of $\Xcal$ is defined as
		\begin{equation*}
		\setlength{\abovedisplayskip}{1mm}
		\setlength{\belowdisplayskip}{1mm}
		\footnotesize
		\begin{aligned}
			\Vcal(\Xcal) := \min_{w\in {\rm span}\rbr{\Dcal}}\max_{x\in\Xcal}\frac{\inner{w}{x}_{\Dcal}}{\nbr{w}_{\Dcal}^2}.
		\end{aligned}
		\end{equation*}
	\end{definition}
	\vspace{-2mm}
	Then the result via exact recovery is given as follows.
	\vspace{-2mm}
	\begin{theorem}\label{thm:pool_exact}
		\xlm{Suppose} that the student answers questions in the exam phase via $\thickmuskip=2mu \medmuskip=2mu F(\cdot) = I(\cdot)$ or $\thickmuskip=2mu \medmuskip=2mu F(\cdot) = S(\cdot)$ and $\thickmuskip=2mu \medmuskip=2mu \Gcal^\top (w^0), v^\ast\in {\rm span}\rbr{\Dcal}$. If $\forall~ \Gcal^\top (w) \in {\rm span}(D)$, there \xlm{exist} $(x, y)\in \Dcal\times\Ycal$ and $\gamma$ such that for $\thickmuskip=2mu \medmuskip=2mu \hat x = \frac{\gamma\nbr{\Gcal^\top (w) - v^\ast}_{\Dcal}}{\nbr{x}_{\Dcal}}x, \, \hat y = y$, we have
		\begin{equation*}
		\setlength{\abovedisplayskip}{-1mm}
		\setlength{\belowdisplayskip}{0mm}
		\footnotesize
		\begin{aligned}
		0\le \gamma \nabla_{\inner{v^t}{\hat x}}\ell\rbr{\inner{v^t}{\hat x}, \hat y}\le \frac{2\Vcal\rbr{\Xcal}\sigma_{\min}}{\eta\sigma^2_{\max}},
		\end{aligned}
		\end{equation*}
		then $(\ell,\Gcal)$ is ET with $\Ocal\rbr{\log\frac{1}{\epsilon}}$ teaching samples and ${\rm rank}(\Dcal)$ query samples.
	\end{theorem}
	\par
\vspace{-2mm}
	For the approximate recovery case, the active learning is no longer able to achieve the desired accuracy for estimating the student's parameter in the restricted pool scenario. Thus the active teacher may not achieve exponential teaching.
\vspace{-3mm}
\section{Discussions and Extensions}
\vspace{-1mm}
\textbf{The active teacher need not know the learning rate.} To estimate the learning rate, the active teacher should first estimate the student's initial parameters $w_1\in R^d$, and then feed the student with one random sample $(x_r, y_r)$. Once the updated student's parameter $w_2$ is estimated by the teacher, the learning rate $\eta$ can be computed by $\thickmuskip=2mu \medmuskip=2mu \eta =\frac{1}{d} \sum \big((w_1 - w_2 ) ./ \nabla_w \ell(w_1^Tx, y)\big)$ where $./$ denotes the element-wise division and the sum is over all the dimensions in $w_1$. The number of samples for estimating $\eta$ will be $2m+1$, where $m$ denotes the samples used in estimating student's parameter. Even if the learning rate is unknown, the teacher only needs $2m+1$ more samples to estimate it. Most importantly, it will not affect the exponential teachability.
\par
\vspace{-1mm}
\textbf{Teaching with forgetting.} We consider the scenario where the learner may forget some knowledge that the teacher has taught, which is very common in human teaching. We model the forgetting behavior of the learner by adding a deviation to the learned parameter. Specifically in one iteration, the learner updates its model with $\thickmuskip=2mu \medmuskip=2mu w^{t+1}=w^t+\nabla_w\ell(\langle w^t,x\rangle,y)$, but due to the forgetting, its truly learned parameter $\hat{w}^{t+1}$ is $\thickmuskip=2mu \medmuskip=2mu w^{t+1}+\epsilon_t$ where $\epsilon_t$ is a random deviation vector. Based on Theorem \ref{theorem:recursion_with_error}, we can show that such forgetting learner is not ET with a teacher that only knows the learner's initial parameter and can not observe the learner along iteration. However, the active teacher can make the forgetting learner ET via the active query strategy. More details and experiments are provided in Appendix \ref{forgetting}.
\par
\vspace{-1mm}
\textbf{Teaching by multiple teachers.} Suppose multiple teachers sequentially teach a learner, a teacher can not guide the learner without knowing its current parameter. It is natural for the teacher to actively estimate the learner. Our active teaching can be easily extended to multiple teacher scenario.

    \vspace{-2mm}
	\section{Experiments}
	\vspace{-1mm}
	
	\textbf{General settings.} Detailed \xlm{settings} are given in Appendix \ref{appendix:exp}. We mainly evaluate the practical pool-based teaching (without rescaling) in the experiments. Still, in the exam stage, our active teacher is able to synthesize novel query examples as needed. The active teacher works in a different feature space from the learner's space, while the omniscient teacher \cite{liu2017iterative} can fully observe the learner and works in the same feature space as the learner. The omniscient teacher serves as a baseline (possibly an upper bound) in our experiments. For active learning, we use the algorithm in \cite{balcan2009agnostic,schein2007active}.
	\par
\vspace{-1mm}
	\textbf{Evaluation.} For synthetic data, we use two metrics to evaluate the convergence performance: \xlm{the objective value and $\thickmuskip=2mu \medmuskip=2mu \nbr{\xlm{\Gcal^\top (w^t)}-v^*}_2$ w.r.t. the training set}. For real images, we further use accuracy on the testing set for evaluation. We put the experiments of forgetting learner in Appendix \ref{forgetting}.
	
	\vspace{-2.3mm}
	\subsection{Teaching with Synthetic Data}
    \vspace{-1.2mm}

\begin{figure}[t]
    	\centering
    	\footnotesize
    	\renewcommand{\captionlabelfont}{\footnotesize}
    	\vspace{-2mm}
    	\begin{tabular}{cc}
    		\includegraphics[width=0.45\linewidth]{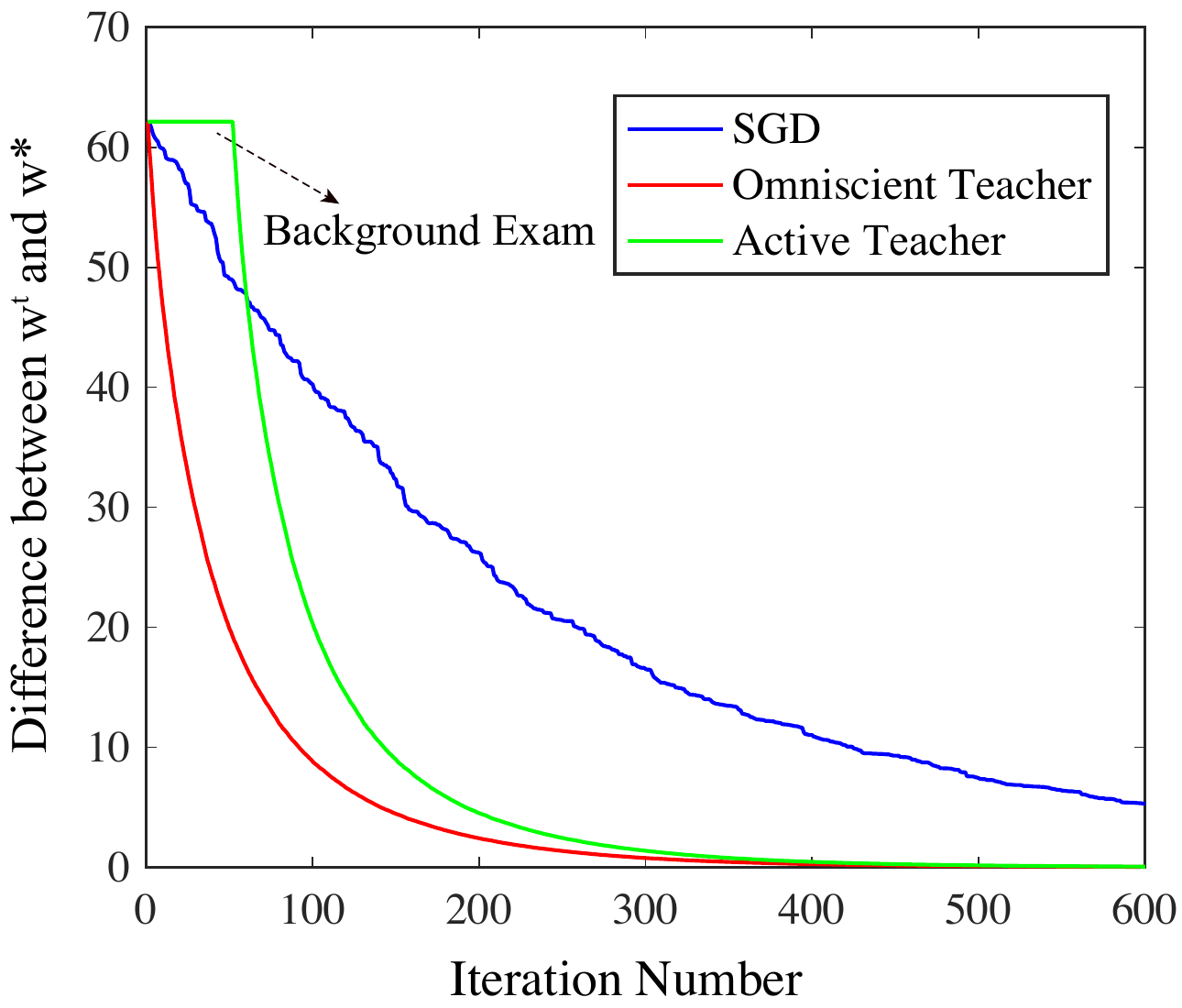} &
    		\hspace{-0.09in}\includegraphics[width=0.46\linewidth]{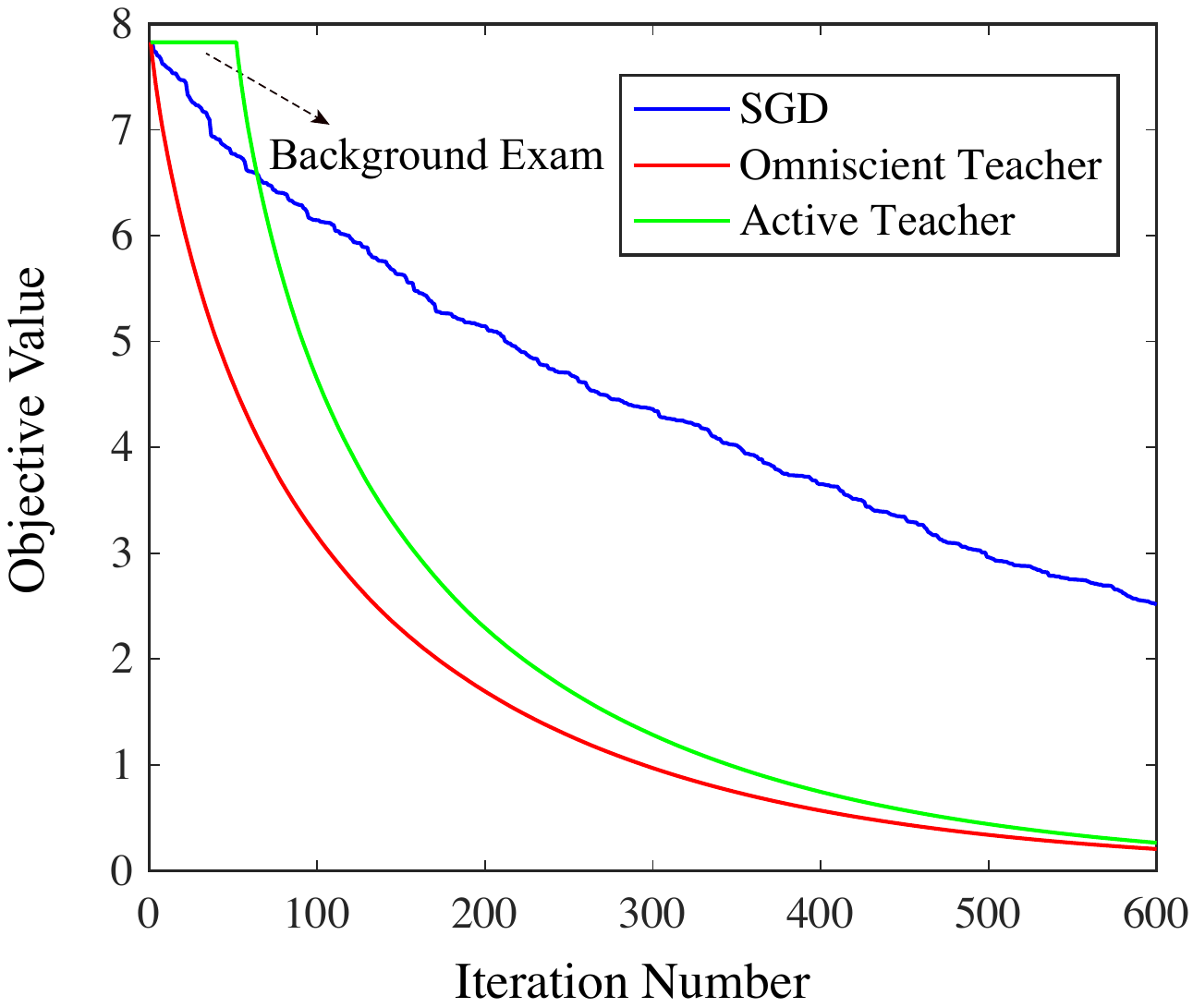}\\
    		LSR & LSR \\
    		\vspace{-0.08in}\\
    		\includegraphics[width=0.45\linewidth]{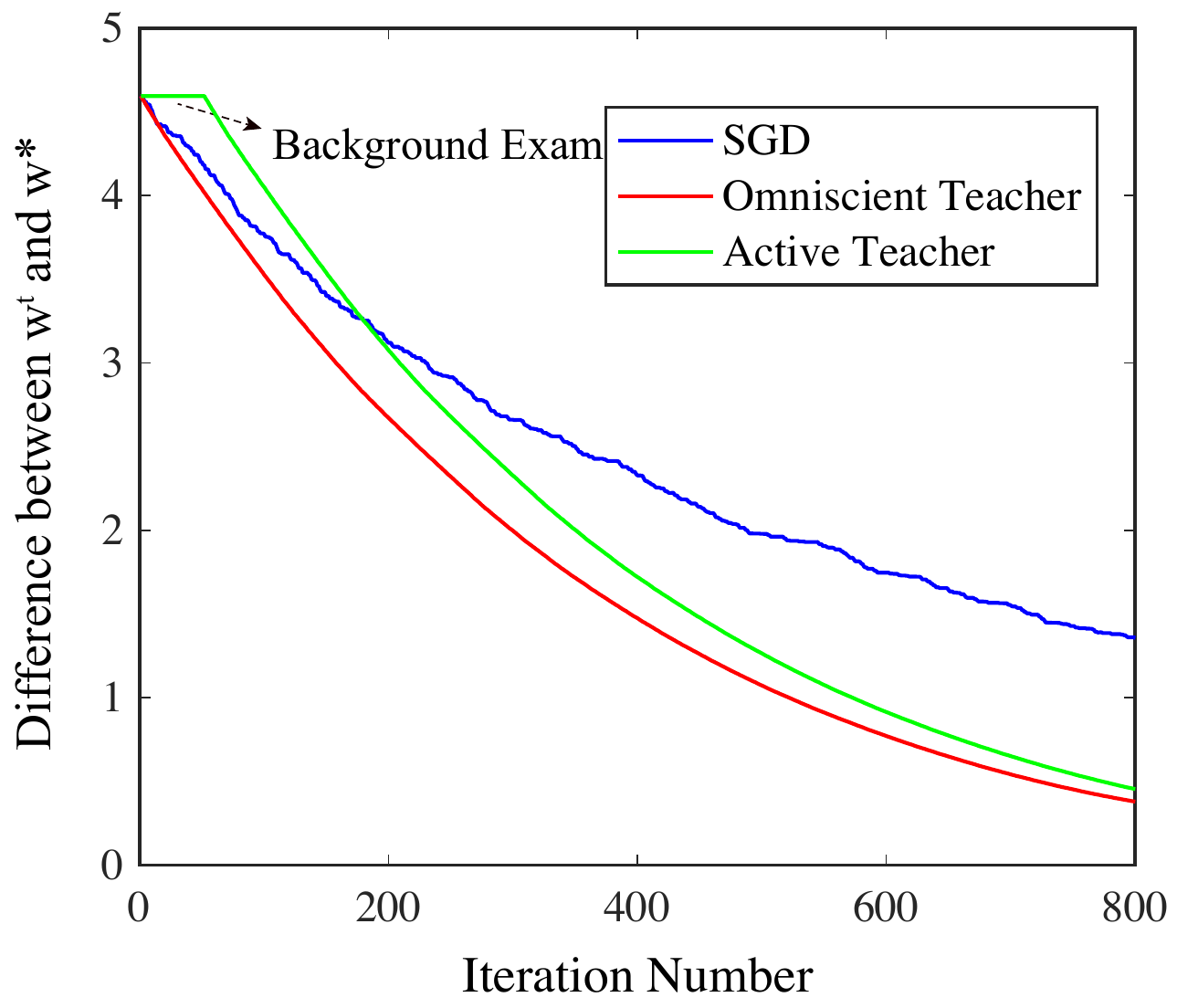} &
    		\hspace{-0.09in}\includegraphics[width=0.45\linewidth]{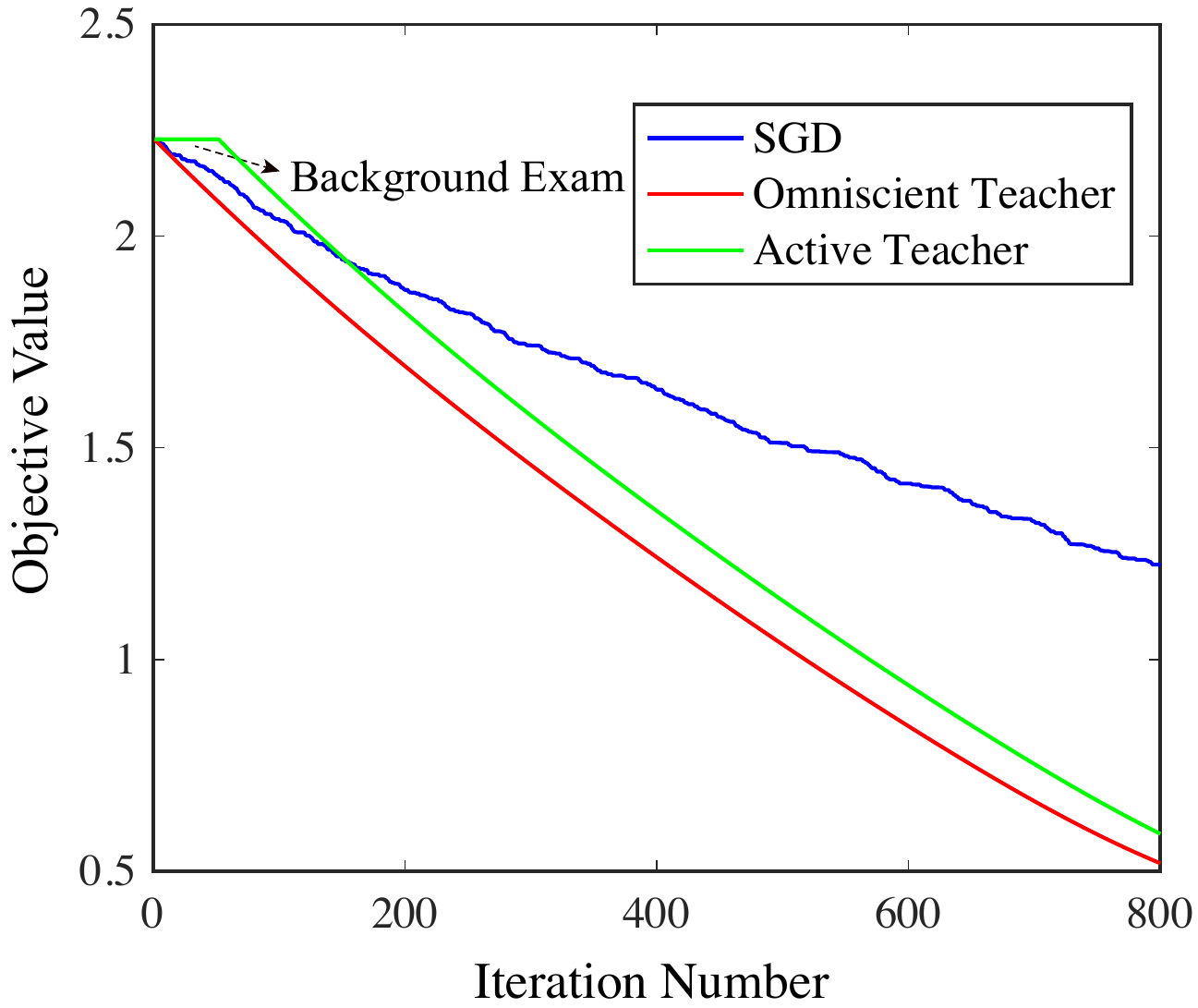}\\
    		LR ($F(z)$ is sigmoid)  & LR ($F(z)$ is sigmoid) \\
    		\vspace{-0.08in}\\
    		\includegraphics[width=0.44\linewidth]{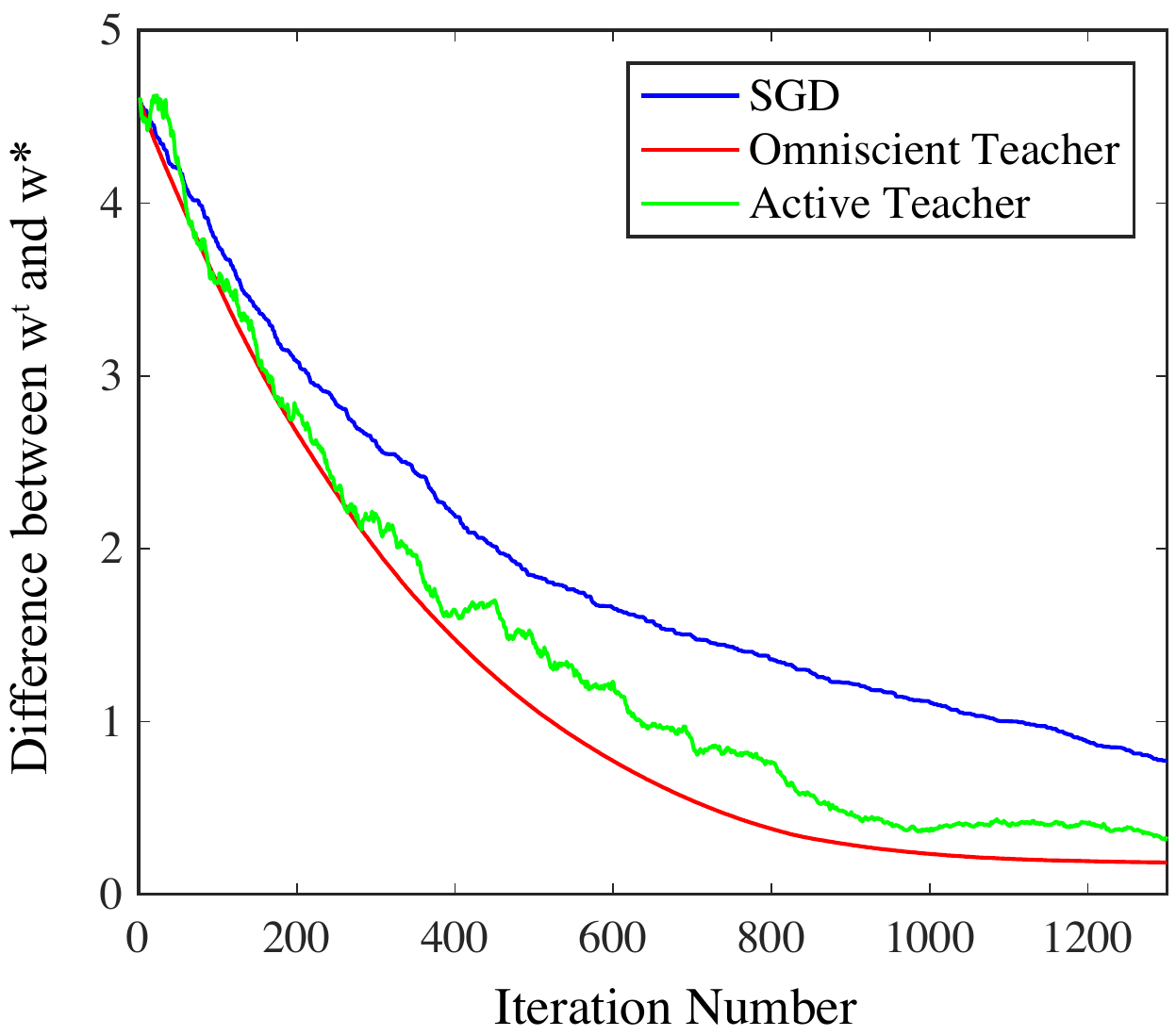} &
    		\hspace{-0.09in}\includegraphics[width=0.44\linewidth]{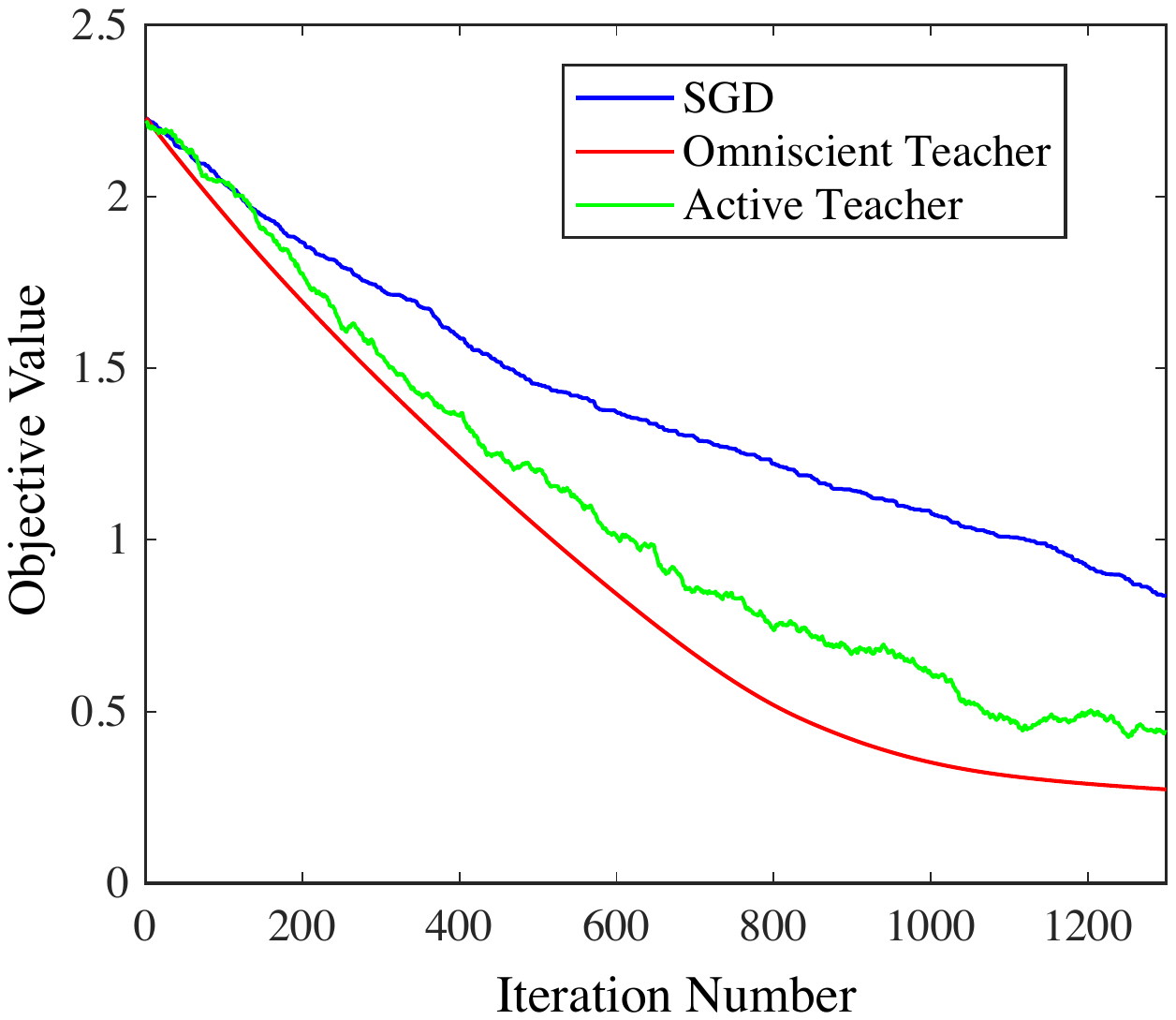}\\
    		LR ($F(z)$ is sign)  & LR ($F(z)$ is sign)
    	\end{tabular}
    	\vspace{-2.5mm}
    	\caption{\footnotesize The convergence performance of random teacher (SGD), omniscient teacher and active teacher. As we need to perform the active query in each iteration for logistic regression ($F(z)$ is sign), we remove the iteration for fair comparison. We only show the teaching complexity for fair comparison.}\label{Gau_data}
    	\vspace{-5mm}
    \end{figure}

    We use Gaussian distributed data to evaluate our active teacher model \xlm{on} linear regression and binary linear classification tasks. We study the LRS learner with $\thickmuskip=2mu \medmuskip=2mu F(\inner{w}{\tilde{x}})=\inner{w}{\tilde{x}}$, LR learner with $\thickmuskip=2mu \medmuskip=2mu F(\inner{w}{\tilde{x}})$\xlm{ being the sigmoid function}, LR learner with $\thickmuskip=2mu \medmuskip=2mu F(\inner{w}{\tilde{x}})=\textnormal{sign}(\inner{w}{\tilde{x}})$. For the first two cases, the active teacher can perform an one-time exam (\xlm{``background exam''}) to exactly recover the ideal virtual learner. After recovering the ideal virtual \xlm{learner}, the active teaching could achieve the performance of the omniscient teaching. The experimental results in Fig. \ref{Gau_data}(a) and Fig. \ref{Gau_data}(b) meet our expectations. In \xlm{the initial iterations} (\xlm{on the order of} feature dimensions), we \xlm{can} see that the learner does not update itself. In this stage, the active teacher provides query samples to the learner and recover a virtual learner based on the feedbacks of these query samples. After the exact recovery of the virtual learner, one can observe that the active teacher achieves faster convergence compared \xlm{with} the random teacher (SGD). In fact, the active teacher and the omniscient teacher should achieve the same convergence speed if omitting numerical errors.
    \par
    \vspace{-0.5mm}
    For the LR learner with $\thickmuskip=2mu \medmuskip=2mu F(\inner{w}{\tilde{x}})=\textnormal{sign}(\inner{w}{\tilde{x}})$, the teacher could only approximate the learner with the active learning algorithm. Besides, the active teacher needs to know the norm of the student model. We use the algorithm in \cite{schein2007active} and recover the virtual learner in each iteration such that $\thickmuskip=2mu \medmuskip=2mu \|\hat{\Gcal}^\top(w)-\Gcal^\top(w)\|_2$ \xlm{becomes} small enough. From the results in Fig. \ref{Gau_data}(c), we \xlm{can} see that due to the approximation error between the recovered virtual learner and the ideal virtual learner, the active teacher can not achieve the same performance as the omniscient teacher. However, the convergence of the \xlm{active teacher} is very close to the omniscient teacher, and is still much faster than SGD. Note that, we remove the iterations used for exams to better compare the convergence of different approaches.

    \vspace{-2.3mm}
	\subsection{Teaching with Real Image Data}\label{mnist_exp_sect}
	\vspace{-1.2mm}
	
	We apply the active teacher to teach the LR learner on the MNIST dataset \cite{lecun1998gradient} to further evaluate the performance. In this experiment, we perform binary classification on the digits 7 and 9. We use two random projections to obtain two sets of 24-dim features for each image: one is for the teacher's feature space and the other is for the student's feature space. The omniscient teacher uses the student's space as its own space (\ie, shared feature space), while the active teacher uses different feature space with the student. For the LR learner with sign function (\ie\ 1-bit feedbacks), one \xlm{can} observe that the active teacher \xlm{has comparable performance} to the omniscient teacher, even doing better at the beginning. Because we evaluate the teaching performance on real image data, the omniscient teacher will not necessarily be an upper bound of all the teacher. Still, as the algorithms \xlm{iterate}, the active teacher becomes worse than the omniscient teacher due to its approximation error.
	\par
\begin{figure}[t]
		%		\vspace{-1.5mm}
		\centering
		\footnotesize
		\renewcommand{\captionlabelfont}{\footnotesize}
		\vspace{-2mm}
		\begin{tabular}{cc}
			\hspace{-0.09in}\includegraphics[width=0.45\linewidth]{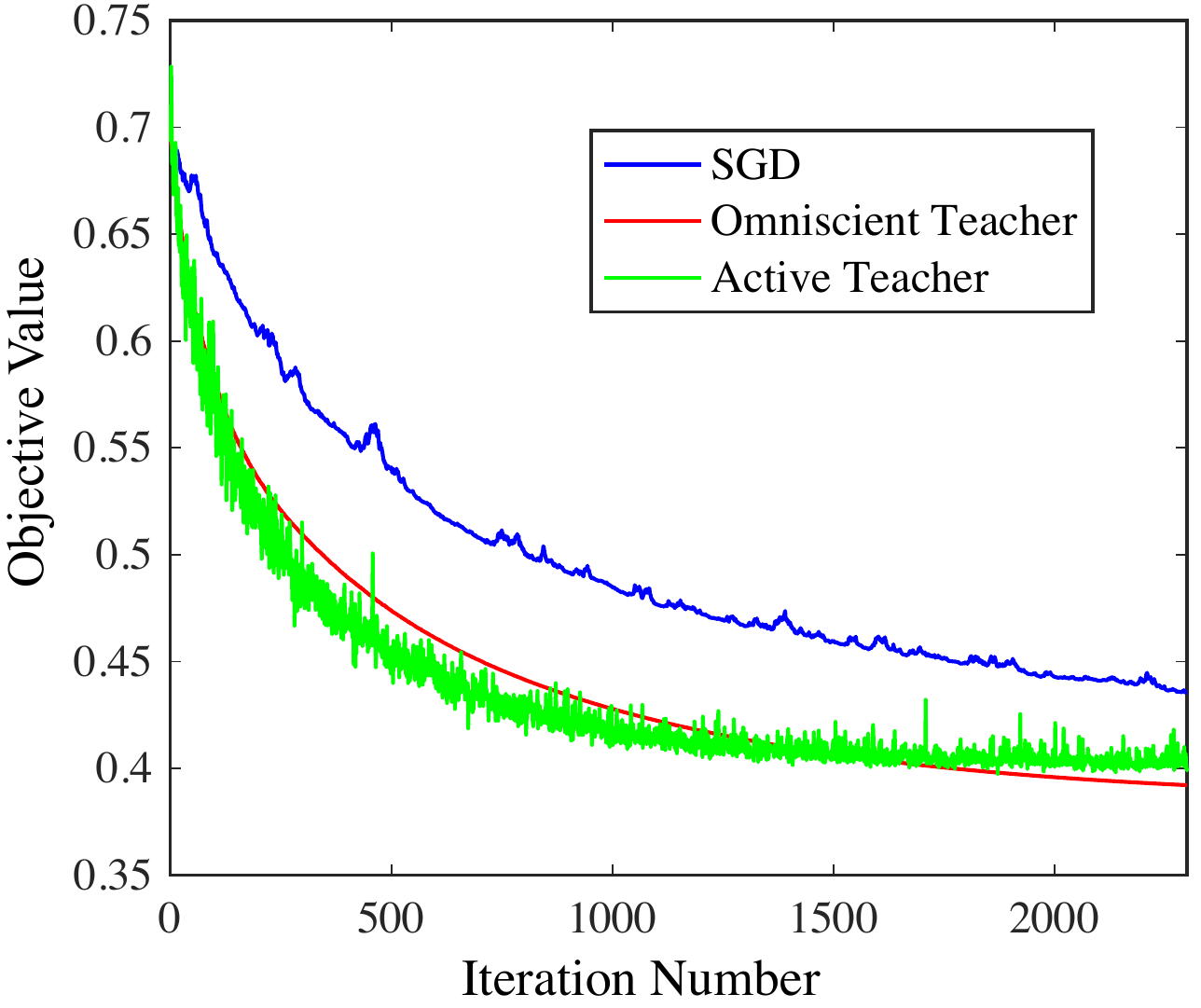} &
			\hspace{-0.11in}\includegraphics[width=0.46\linewidth]{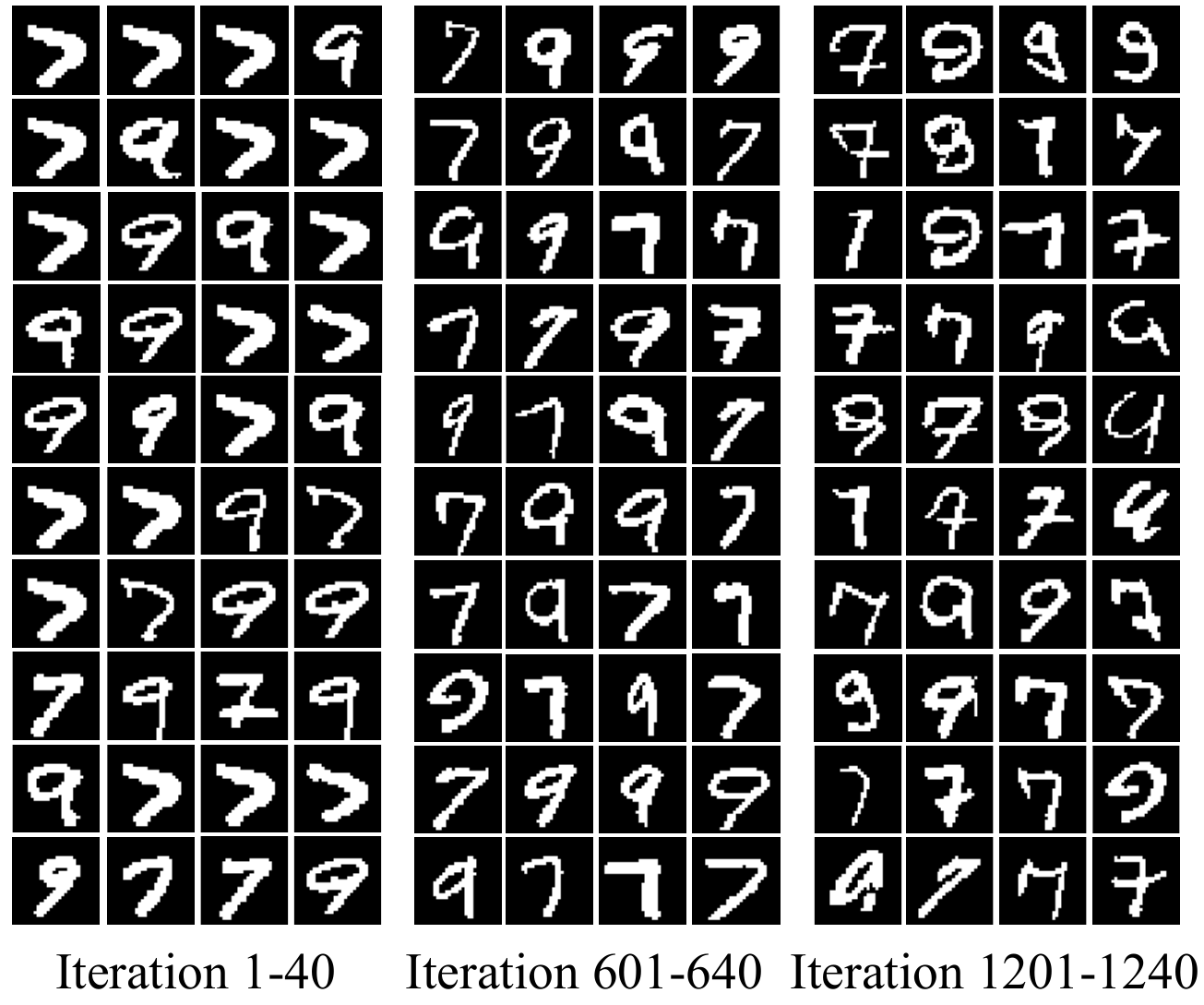}\\
			LR ($F(z)$ is sign) & Active Teacher \\
			\vspace{-0.08in}\\
			\hspace{-0.09in}\includegraphics[width=0.45\linewidth]{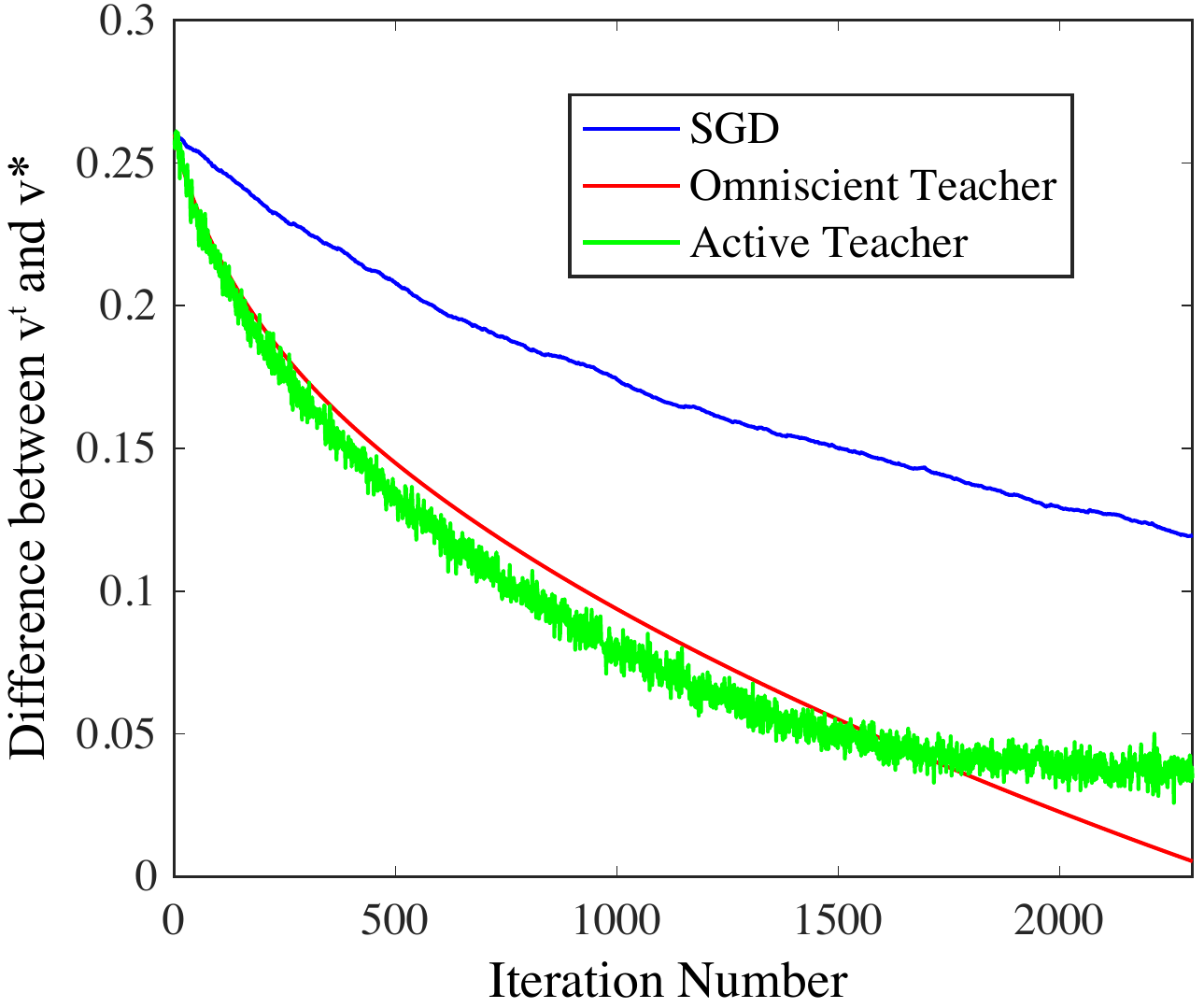} &
			\hspace{-0.11in}\includegraphics[width=0.46\linewidth]{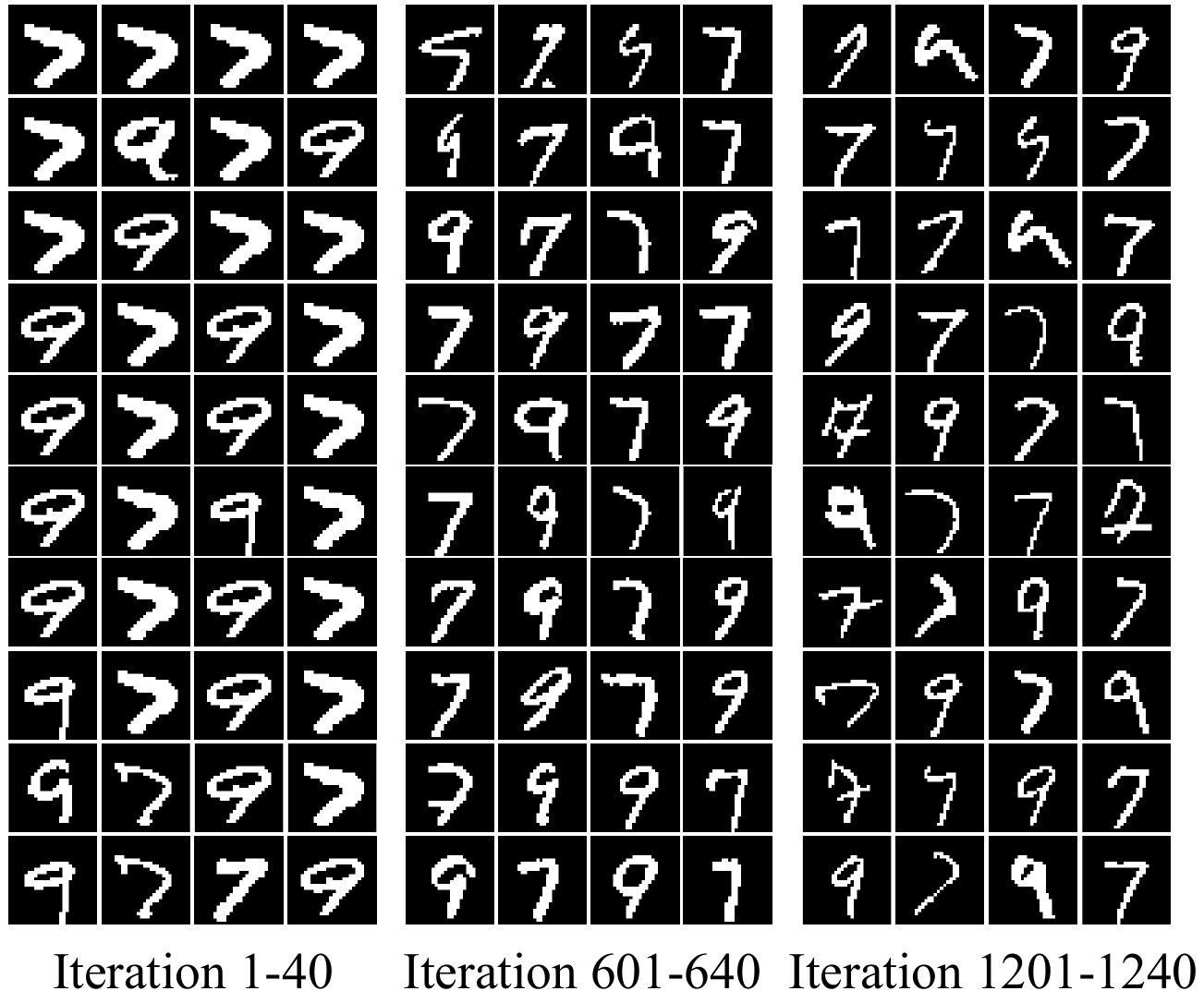}\\
			LR ($F(z)$ is sign)  & Omniscient Teacher \\
			\vspace{-0.08in}\\
			\hspace{-0.06in}\includegraphics[width=0.46\linewidth]{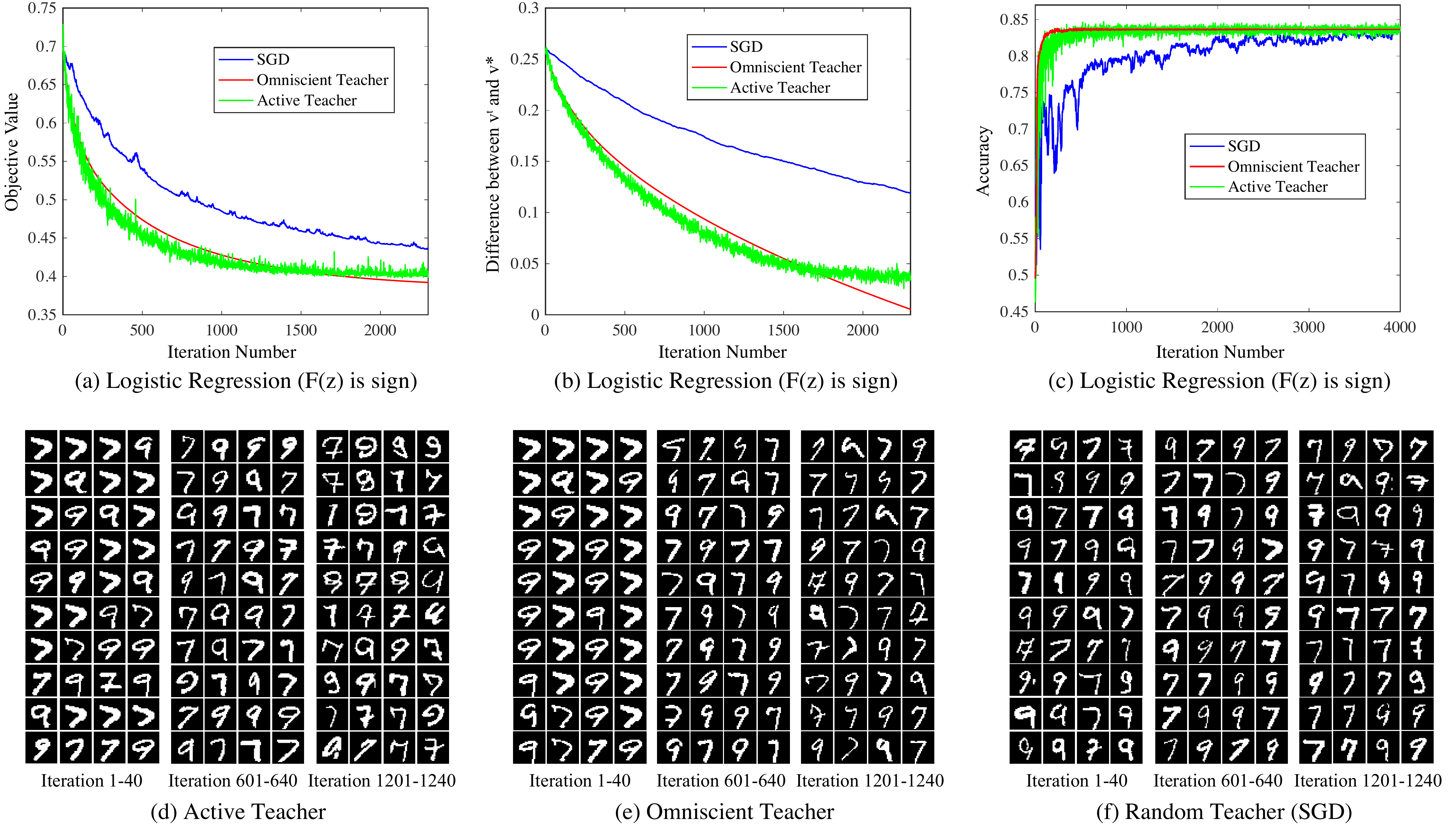} &
			\hspace{-0.15in}\includegraphics[width=0.46\linewidth]{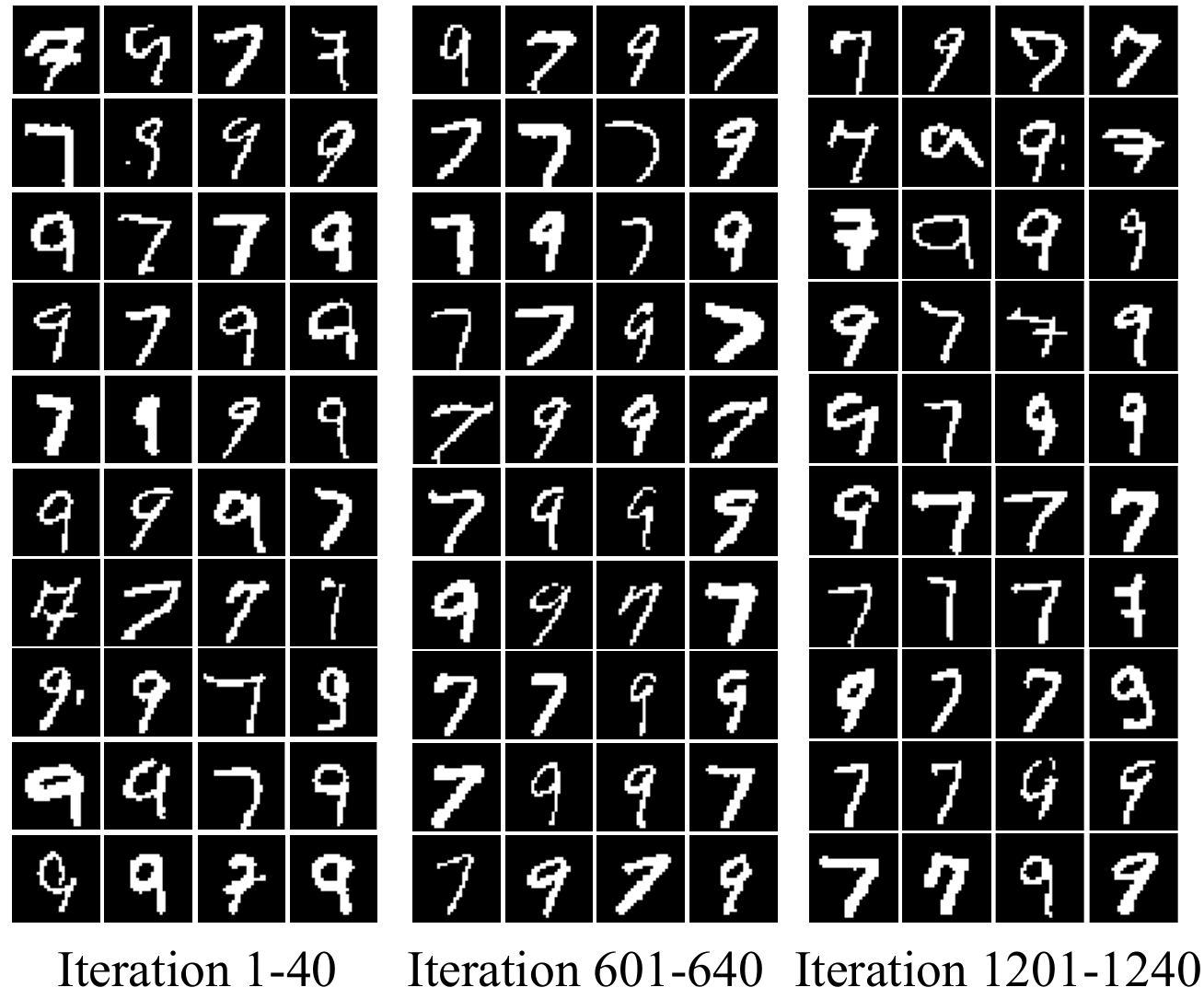}\\
			LR ($F(z)$ is sign)  & Random Teacher (SGD)
		\end{tabular}
		\vspace{-2mm}
		\caption{\footnotesize The convergence performance of random teacher (SGD), omniscient teacher and active teacher in MNIST 7/9 classification. Similar to the previous, we only show the teaching complexity for fair comparison. More experiments on the logistic regression with $\thickmuskip=2mu \medmuskip=2mu F(z)=S(z)$ is in Appendix \ref{appendix:more}.}\label{mnist_data}
		\vspace{-5mm}
	\end{figure}

\vspace{-1mm}
	In the right side of Fig.\ref{mnist_data}, we visualize the images selected by the active teacher, omniscient teacher and random teacher. The active teacher preserves the pattern of images selected by the omniscient teacher: starting from easy examples first and gradually shifting to difficult ones, while the images selected by the random teacher have no patterns.

	\vspace{-3.5mm}
	\section{Conclusions and Open Problems}
	\vspace{-1.5mm}
	
	As a step towards the ultimate black-box machine teaching, cross-space teaching greatly relaxes the assumptions of previous teaching scenarios and bridges the gap between the iterative machine teaching and the practical world. The active teaching strategy is inspired by realistic human teaching. For machine teaching to be applicable in practice, we need to gradually remove all the unrealistic assumptions to obtain more realistic teaching scenario. The benefits of more realistic machine teaching are in two folds. First, it enables us make better use of the existing off-the-shelf pretrained models to teach a new model on some new tasks. It is also related to transfer learning~\cite{pan2010survey}. Second, it can improve our understanding on human education and provide more effective teaching strategies for humans.
	\par
	\vspace{-0.5mm}
	\textbf{Rescalable pool-based active teaching with $1$-bit feedback.} The proposed algorithm may not work the in pool-based teaching setting when the student return $1$-bit feedback. We leave the possibility of achieving exponential teachability in \xlm{this} setting as an open problem.
	\par
	\vspace{-0.5mm}
	\textbf{Relaxation for the conditions on $\Gcal$.} Current constraints on the operator $\Gcal$ are still too strong to match more practical scenarios. How to relax the conditions on $\Gcal$ is important.
	\par
	\vspace{-0.5mm}
	\textbf{A better alternative to approximate recovery?} Is there some other tool other than active learning for our teacher to recover the virtual learner? For example, 1-bit compressive sensing \cite{boufounos20081} \xlm{may help}.
	
	\newpage
	\section*{Acknowledgements}
	The project was supported in part by NSF IIS-1218749, NSF Award BCS-1524565, NIH BIGDATA 1R01GM108341, NSF CAREER IIS-1350983, NSF IIS-1639792 EAGER, NSF CNS-1704701, ONR N00014-15-1-2340, Intel ISTC, NVIDIA, and Amazon AWS.
	\bibliographystyle{icml2018}
	{\small \bibliography{active_teaching}}

%%%%%%%%%%%%%%%%%%%%%%%%%%%%%%%%%%%%%%%%%%%%%%%%%%%%%%%%%%%%%%%%%%%%%%%%%%%%%%%%%%%%%%%%%%%%%%%%%%%
% Appendix
%%%%%%%%%%%%%%%%%%%%%%%%%%%%%%%%%%%%%%%%%%%%%%%%%%%%%%%%%%%%%%%%%%%%%%%%%%%%%%%%%%%%%%%%%%%%%%%%%%%
\clearpage
\newpage

\appendix
\onecolumn

\begin{appendix}
	
\begin{center}
	{\Large \bf Appendix}
\end{center}

%%%------------------------------------------------------------------------------------------------
\section{Details of the Proofs}
%%%------------------------------------------------------------------------------------------------

We analyze the sample complexity by \xlm{separating} the teaching procedure into two stages in each iteration, \ie, \xlm{the }active query stage by conducting examination for the student and \xlm{the} teaching stage by providing samples to the student.

%%%------------------------------------------------------------------------------------------------
\subsection{Error Decomposition}
%%%------------------------------------------------------------------------------------------------
Recall \xlm{that} there is a mapping $\Gcal$ from the feature space \xlm{of} the teacher to that of \xlm{the} student, \xlm{and} we have $\inner{w}{\xtil} = \inner{w}{\Gcal(x)} = \inner{\Gcal^\top(w)}{x}$ where $\Gcal^\top$ denotes the conjugate mapping  of $\Gcal$. We also denote the $\sigma_{\max} = \max_{x^\top x=1} \Gcal^\top(x) \Gcal(x)$, $\sigma_{\min} = \min_{x^\top x=1} \Gcal^\top(x) \Gcal(x) >0 $ since the operator $\Gcal$ is invertible, and $\kappa\rbr{\Gcal^\top\Gcal} = \frac{\sigma_{\max}}{\sigma_{\min}}$. To involve the inconsistency between the student's parameters $w^{t}$, and the teacher's estimator $v^t$, \xlm{at} $t$-th iteration into the analysis, we first provide the recurrsion with error decomposition. For simplicity, we denote $\beta{\rbr{\inner{w}{x}, y}} := \nabla_{\inner{w}{x}}\ell\rbr{\inner{w}{x}, y}$.
Then, we have the update rule of student as
$$
w^{t+1} = w^t - \eta\beta{\rbr{\inner{w^t}{\Gcal(x^t)}, y^t}}\Gcal(x^t),
$$
where $x^t = \gamma\rbr{v^t - v^\ast}$ is constructed by teacher with the estimator $v^t$. Plug into the difference, we have
\begin{align*}
	&\nbr{\xlm{\Gcal^\top (w^{t+1})} - v^\ast}^2 \\
	&= \nbr{\xlm{\Gcal^\top (w^t)} - v^\ast}^2 + \eta^2\beta^2\rbr{\inner{w^t}{\Gcal(x^t)}, y^t}\nbr{\Gcal^\top\Gcal (x^t)}^2 - 2\eta\beta\rbr{\inner{w^t}{\Gcal(x^t)}, y^t}\inner{\Gcal^\top\Gcal (x^t)}{\xlm{\Gcal^\top (w^t)} - v^\ast}\\
	&=\nbr{\xlm{\Gcal^\top (w^t)} - v^\ast}^2 + \eta^2\beta^2\rbr{\inner{v^t}{x^t}, y^t}\nbr{\Gcal^\top\Gcal (x^t)}^2 - 2\eta\beta\rbr{\inner{v^t}{x^t}, y^t}\inner{\Gcal^\top\Gcal (x^t)}{\xlm{\Gcal^\top (w^t)} - v^\ast}\\
	&\hspace{0.5in} + \eta^2\nbr{\Gcal^\top\Gcal (x^t)}^2\rbr{\beta^2\rbr{\inner{\xlm{\Gcal^\top (w^t)}}{x^t}, y^t} - \beta^2\rbr{\inner{v^t}{x^t}, y^t}}\\
	&\hspace{0.5in} - 2\eta\inner{\Gcal^\top\Gcal (x^t)}{\xlm{\Gcal^\top (w^t)} - v^\ast}\rbr{\beta\rbr{\inner{\xlm{\Gcal^\top (w^t)}}{x^t}, y^t} - \beta\rbr{\inner{v^t}{x^t},y^t}}.
\end{align*}
Suppose the loss function is \xlm{$L$}-Lipschitz smooth and $x\in \Xcal = \cbr{x\in \RR^d, \nbr{x}\le R}$,
$$
\xlm{\abr{\beta\rbr{\inner{v_1}{x}, y} - \beta\rbr{\inner{v_2}{x}, y}}} \le LR\nbr{v_1 - v_2},
$$
which implies
$$
\beta\rbr{\inner{v_2}{x}, y} - LR\nbr{v_1 - v_2} \le \beta\rbr{\inner{v_1}{x}, y} \le \beta\rbr{\inner{v_2}{x}, y} + LR\nbr{v_1 - v_2}.
$$

We have the error decomposition as follows,
\begin{align}\label{eq:error_decomposition}
\nbr{\xlm{\Gcal^\top (w^{t+1})} - v^\ast}^2 &\le \nbr{\xlm{\Gcal^\top (w^t)} - v^\ast}^2 + \eta^2\beta^2\rbr{\inner{v^t}{\gamma(v^t - v^\ast)}, y^t}\gamma^2\nbr{\Gcal^\top\Gcal (v^t - v^\ast)}^2  \nonumber\\
&\hspace{0.3in}- 2\eta\beta\rbr{\inner{v^t}{\gamma(v^t - v^\ast)}, y^t}\gamma\inner{\Gcal^\top\Gcal (v^t - v^\ast)}{\xlm{\Gcal^\top (w^t)} - v^\ast}\nonumber\\
&\hspace{0.3in} + \eta^2\gamma^2LR\nbr{\Gcal^\top\Gcal (v^t - v^\ast)}^2\nbr{\xlm{\Gcal^\top (w^t)} - v^t}\rbr{\beta\rbr{\inner{\xlm{\Gcal^\top (w^t)}}{x^t}, y^t} + \beta\rbr{\inner{v^t}{x^t}, y^t}}\nonumber\\
&\hspace{0.3in} + 2\eta \gamma LR\inner{\Gcal^\top\Gcal (v^t - v^\ast)}{\xlm{\Gcal^\top (w^t)} - v^\ast}\nbr{\xlm{\Gcal^\top (w^t)} - v^t}\nonumber\\
%%%%%%%%%%%%%%
&\le\nbr{\xlm{\Gcal^\top (w^t)} - v^\ast}^2 + \eta^2\gamma^2\sigma_{\max}^2{\beta^2\rbr{\inner{v^t}{\gamma(v^t - v^\ast), y^t}}}\nbr{v^t - v^\ast}^2 \\
&\hspace{0.3in}- 2\eta\beta\rbr{\inner{v^t}{\gamma(v^t - v^\ast)}, y^t}\gamma\rbr{\sigma_{\min}\nbr{v^t - v^\ast}^2 - \sigma_{\max}\nbr{\xlm{\Gcal^\top (w^t)} - v^t}\nbr{v^t - v^\ast}} \nonumber\\
&\hspace{0.3in} + \eta^2\gamma^2LR\nbr{\Gcal^\top\Gcal (v^t - v^\ast)}^2\nbr{\xlm{\Gcal^\top (w^t)} - v^t}\rbr{{2\beta\rbr{\inner{v^t}{x^t}, y^t}}  + LR \nbr{\xlm{\Gcal^\top (w^t)} - v^t} }\nonumber\\
&\hspace{0.3in} + 2\eta \gamma LR\rbr{\nbr{\Gcal^\top\Gcal}\nbr{v^t - v^\ast}^2 + \nbr{\Gcal^\top\Gcal}\nbr{v^t - v^\ast}\nbr{\xlm{\Gcal^\top (w^t)} - v^t}}\nbr{\xlm{\Gcal^\top (w^t)} - v^t}\nonumber
\end{align}
where the last two terms represent the inconsistency \xlm{on the teacher's side and the student's side} in computing $\beta$.

%%%------------------------------------------------------------------------------------------------
\subsection{Exact Recovery of $\xlm{\Gcal^\top (w)}$}
%%%------------------------------------------------------------------------------------------------

{\bf Theorem~\ref{thm:exact_recovery}}
\emph{
\xlm{Suppose the teacher is able to recover $\Gcal^\top (w^t)$ exactly using $m$ samples at each iteration. If for any $\thickmuskip=2mu \medmuskip=2mu v\in\RR^d$, there exists $\thickmuskip=2mu \medmuskip=2mu \gamma\neq 0$ and $\hat y$ such that $\thickmuskip=2mu \medmuskip=2mu \hat x = \gamma\rbr{v - v^*}$ and}
\begin{align*}
0 < \gamma\nabla_{\inner{v^t}{\hat x}}\ell\rbr{\inner{v^t}{\hat x}, \hat y} < \frac{2\sigma_{\min}}{\eta\sigma^2_{\max}},
\end{align*}
then $(\ell,\Gcal)$ is ET with $\Ocal\rbr{\rbr{m+1}\log\frac{1}{\epsilon}}$ samples.
}
\begin{proof}
	Plug $\nbr{\xlm{\Gcal^\top (w^t)} - v^t} = 0$ into the error decomposition~\eq{eq:error_decomposition}, we have
	\begin{eqnarray*}
		\nbr{\xlm{\Gcal^\top (w^{t+1})} - v^\ast}^2	&\le&\nbr{\xlm{\Gcal^\top (w^t)} - v^\ast}^2 + \eta^2\gamma^2\sigma_{\max}^2{\beta^2\rbr{\inner{v^t}{\gamma(v^t - v^\ast), y^t}}}\nbr{v^t - v^\ast}^2 \\
		&&- 2\eta\beta\rbr{\inner{v^t}{\gamma(v^t - v^\ast)}, y^t}\gamma\rbr{\sigma_{\min}\nbr{v^t - v^\ast}^2}\\
		&\le&\rbr{1 + \eta^2\gamma^2\sigma_{\max}^2{\beta^2\rbr{\inner{v^t}{\gamma(v^t - v^\ast), y^t}}} - 2\eta\beta\rbr{\inner{v^t}{\gamma(v^t - v^\ast)}, y^t}\gamma{\sigma_{\min}}}	\nbr{\xlm{\Gcal^\top (w^t)} - v^\ast}^2.
	\end{eqnarray*}
	Denote $\nu\rbr{\gamma} = \min_{\hat x\in \Xcal, \hat y\in \Ycal} \gamma \beta\rbr{\inner{v^t}{\gamma(v^t - v^\ast)}, y^t} >0$, and $\mu\rbr{\gamma} = \max_{\hat x\in \Xcal, \hat y\in \Ycal} \gamma \beta\rbr{\inner{v^t}{\gamma(v^t - v^\ast)}, y^t} <\frac{2\sigma_{\min}}{\eta\sigma^2_{\max}}$, we have the recursion
	$$
	\nbr{\xlm{\Gcal^\top (w^{t+1})} - v^\ast}^2\le r\rbr{\eta, \gamma} \nbr{\xlm{\Gcal^\top (w^{t})} - v^\ast}^2,
	$$
	where $r\rbr{\eta, \gamma} = \max\cbr{1 + \eta^2\sigma^2_{\max}\mu\rbr{\gamma} - 2\eta\sigma_{\min}\mu\rbr{\gamma}, 1 + \eta^2\sigma^2_{\max}\nu\rbr{\gamma} - 2\eta\sigma_{\min}\nu\rbr{\gamma}}$ and $0 \le r\rbr{\eta, \gamma}\le 1$. Therefore, the algorithm converges exponentially,
	$$
	\nbr{\xlm{\Gcal^\top (w^t)}-v^\ast}\le r\rbr{\eta, \gamma}^{t/2}\nbr{\xlm{\Gcal^\top (w^0)} - v^\ast}.
	$$
	In other words, the students needs $2\rbr{\log\frac{1}{r\rbr{\eta, \gamma}}}^{-1}\log\frac{\nbr{\xlm{\Gcal^\top (w^0)} - v^*}}{\epsilon}$ samples for updating. \xlm{Consider that at} each iteration, \xlm{if} the teacher first \xlm{uses} $m$ samples for estimating $\xlm{\Gcal^\top (w)}$, \xlm{then} the total \xlm{number of samples is no larger than} $\rbr{m+1}2\rbr{\log\frac{1}{r\rbr{\eta, \gamma}}}^{-1}\log\frac{\nbr{\xlm{\Gcal^\top (w^0)} - v^*}}{\epsilon}$.
\end{proof}

{\bf Lemma~\ref{thm:invertible_F}}
\emph{
If \xlm{$F(\cdot)$ is} bijective, then we can exactly recover $\thickmuskip=2mu \medmuskip=2mu \Gcal^\top (w)\in\RR^d$ with $d$ samples.
}

\begin{proof}
We prove the theorem by construction. Denote $d$ independent samples as $Z = \cbr{z_i}_{i=1}^d \in \RR^d$. We can exactly recover arbitrary $v$ with these samples by solving the linear system,
\begin{equation}\label{eq:exact_ls}
\inner{v}{Z} = b,
\end{equation}
where $b = F^{-1}\rbr{F\rbr{\inner{w}{\xlm{\Gcal (x)}}}}$ are provided by the student. $F^{-1}$ exists because $F$ is bijective. Since ${\rm rank}(Z) = d$, the linear system~\eq{eq:exact_ls} has \xlm{a} unique solution.
\end{proof}
{\bf Lemma~\ref{thm:hinge_F}}
\emph{
If $F(\cdot) = \max\rbr{0, \cdot}$, then we can exactly recover $\Gcal^\top (w)\in \RR^d$ with $2d$ samples.
}
\begin{proof}
We prove the \xlm{lemma} by construction. Notice that $\forall a\in \xlm{\RR}$, either $\max\rbr{0, a} = a$ and $\max\rbr{0, -a} = 0$, or $\max\rbr{0, a} = 0$ and $\max\rbr{0, -a} = -a$. Then, we can first construct $d$ independent samples as $\cbr{z_i}_{i=1}^d \in \RR^d$, and then, extend the set with $\cbr{-z_i}_{i=1}^d$. We construct the linear system by picking one of the linear equations from $\inner{v}{z_i} = \max\rbr{0, \inner{w}{\Gcal(z_i)}}$ or $\inner{v}{-z_i} = \max\rbr{0, -\inner{w}{\Gcal(z_i)}}$ which does not equal to zero. Denote the linear system $\inner{v}{Z'} = b$, since we select either $z_i$ or $-z_i$ to form $Z$, then, ${\rm rank}(Z') = d$, therefore, the linear system has \xlm{a} unique solution.
\end{proof}

In both regression and classification scenarios, if the student answers the questions in the query phase with $F(\cdot) = I(\cdot)$, $F(\cdot) = S(\cdot)$, or $F(\cdot) = \max\rbr{0, \cdot}$, where $I$ denotes the identity mapping and $S$ denotes some sigmoid function, \eg, logistic function, hyperbolic tangent, error function and so on, we can exactly recover $v = \xlm{\Gcal^\top (w)}\in\RR^d$ with arbitrary $\Ocal(d)$ independent data, omitting the numerical error and consider the solution as exact recovery. Recall we can reuse these $\Ocal(d)$ independent data in each iteration, we have

{\bf Corollary~\ref{cor:identity_sigmoid_F}}
\emph{
Suppose the student answers questions in query phase via $F(\cdot) = I(\cdot)$, $F(\cdot) = S(\cdot)$, or $F(\cdot) = \max\rbr{0, \cdot}$, then $(\ell,\Gcal)$ is ET with $\Ocal\rbr{\log\frac{1}{\epsilon}}$ teaching samples and $\Ocal(d)$ query samples via exact recovery.
}

%%%------------------------------------------------------------------------------------------------
\subsection{Approximate Recovery of $\xlm{\Gcal^\top (w)}$}\label{appendix:approx_recovery}
%%%------------------------------------------------------------------------------------------------
{\bf Theorem~\ref{theorem:recursion_with_error}}
\emph{
Suppose the loss function $\ell$ is $L$-Lipschitz smooth in a compact domain $\thickmuskip=2mu \medmuskip=2mu \Omega_v\subset\RR^d$ of $v$ containing $v^\ast$ and sample candidates $(x, y)$ are from bounded $\thickmuskip=2mu \medmuskip=2mu \Xcal\times\Ycal$, where $\thickmuskip=2mu \medmuskip=2mu \Xcal = \cbr{x\in \RR^d, \nbr{x}\le R}$. Further suppose at $t$-th iteration, the teacher estimates the student $\thickmuskip=2mu \medmuskip=2mu \epsilon_{\text{est}} := \nbr{\Gcal^\top (w^t) - v^t} = \Ocal\rbr{\epsilon}$ with probability at least $1-\delta$ using $m\rbr{\epsilon_{\text{est}}, \delta}$ samples. If for any $v\in\Omega_v$, there exists $\thickmuskip=2mu \medmuskip=2mu \gamma\neq 0$ and $\hat y$ such that for $\thickmuskip=2mu \medmuskip=2mu \hat x = \gamma\rbr{v - v^\ast}$, we have
\begin{align*}
&0 < \gamma\nabla_{\inner{v^t}{\hat x}}\ell\rbr{\inner{v^t}{\hat x}, \hat y} < \frac{2\rbr{1 - \lambda}\sigma_{\min}}{\eta\sigma^2_{\max}}, \quad \\[-1.8mm]
&\text{with  }\, 0<\lambda< \min\big(\frac{\kappa\rbr{\Gcal^\top\Gcal}}{\sqrt{2}},\, 1\big),
\end{align*}
then the student can achieve $\epsilon$-approximation of $v^\ast$ with $\Ocal\rbr{\log\frac{1}{\epsilon}\rbr{1 + m\rbr{\lambda{\epsilon}, \frac{\delta}{\log\frac{1}{\epsilon}}}}}$ samples with probability at least $1-\delta$. If $m\rbr{\epsilon_{\text{est}}, \delta} = \Ocal(\log \frac{1}{\epsilon})$, then $(\ell,\Gcal)$ is ET.
}
\begin{proof}
	Assume that in each iteration, the teacher will estimate the $w^t$ at least satisfying $\epsilon_{\text{est}} := \nbr{\xlm{\Gcal^\top (w^t)} - v^t}\le \lambda\frac{\sigma_{\min}}{\sigma_{\max}}\nbr{v^t - v^\ast}$. Plugging into the error decomposition~\eq{eq:error_decomposition}, we obtain
	\begin{align*}
		\nbr{\xlm{\Gcal^\top (w^{t+1})} - v^\ast}^2 &\le \nbr{\xlm{\Gcal^\top (w^t)} - v^\ast}^2 + \eta^2\gamma^2\sigma_{\max}^2{\beta^2\rbr{\inner{v^t}{\gamma(v^t - v^\ast), y^t}}}\nbr{v^t - v^\ast}^2\\
		& \hspace{0.3in} - 2\eta\beta\rbr{\inner{v^t}{\gamma_t(v^t - v^\ast)}, y^t}\gamma\sigma_{\min} \rbr{1 - \lambda}{\nbr{v^t - v^\ast}^2}\\
		& \hspace{0.3in} + \eta^2LR\epsilon_{\text{est}}\sigma^2_{\max}\gamma^2\nbr{(v^t - v^\ast)}^2\rbr{{2\beta\rbr{\inner{v^t}{x^t}, y^t}}  + LR \epsilon_{\text{est}} }\\
		& \hspace{0.3in} + 2\eta  LR\epsilon_{\text{est}}\rbr{\gamma\sigma_{\max}\nbr{v^t - v^\ast}^2 + \gamma \sigma_{\max}\nbr{\xlm{\Gcal^\top (w^t)} - v^t}\nbr{v^t - v^\ast}}\\
		&\le\nbr{\xlm{\Gcal^\top (w^t)} - v^\ast}^2 + \eta^2\gamma^2\sigma_{\max}^2{\beta^2\rbr{\inner{v^t}{\gamma(v^t - v^\ast), y^t}}}\nbr{v^t - v^\ast}^2\\
		& \hspace{0.3in} - 2\eta\beta\rbr{\inner{v^t}{\gamma_t(v^t - v^\ast)}, y^t}\gamma\sigma_{\min} \rbr{1 - \lambda}{\nbr{v^t - v^\ast}^2}\\
		& \hspace{0.3in} + \eta^2LR^3\epsilon_{\text{est}}\sigma^2_{\max}\rbr{{2\beta\rbr{\inner{v^t}{x^t}, y^t}}  + LR \epsilon_{\text{est}} }\\
		& \hspace{0.3in} + 2\eta  LR^2\epsilon_{\text{est}}\rbr{\sigma_{\max}\nbr{v^t - v^\ast} + \sigma_{\max}\nbr{\xlm{\Gcal^\top (w^t)} - v^t}}
	\end{align*}
	The last inequality due to the fact that $x^t = \gamma\rbr{v^t - v^*}\in \Xcal$, implying $\gamma\nbr{v^t - v^*}\le R$. On the other hand, we have
	\begin{eqnarray*}
		&&\nbr{v^t - v^\ast}^2 = \nbr{v^t - \xlm{\Gcal^\top (w^t)} + \xlm{\Gcal^\top (w^t)} - v^\ast}^2\le 2\nbr{\xlm{\Gcal^\top (w^t)} - v^t}^2 + 2\nbr{\xlm{\Gcal^\top (w^t)}- v^\ast}^2 \\
		&&\le 2\lambda^2\frac{\sigma^2_{\min}}{\sigma^2_{\max}}\nbr{v^t - v^\ast}^2 + 2\nbr{\xlm{\Gcal^\top (w^t)}- v^\ast}^2 \\
		&&\Rightarrow \nbr{v^t - v^\ast}^2 \le \frac{2}{1 - 2\lambda^2\frac{\sigma^2_{\min}}{\sigma^2_{\max}}} \nbr{\xlm{\Gcal^\top (w^t)} - v^\ast}^2.
	\end{eqnarray*}
	Combine this into the recursion,
	\begin{eqnarray}\label{eq:recursion}
	\nbr{\xlm{\Gcal^\top (w^{t+1})} - v^\ast}^2
	\le C_0\nbr{\xlm{\Gcal^\top (w^t)} - v^\ast}^2+ C_1\rbr{\beta\rbr{\inner{v^t}{x^t}, y^t}  + \nbr{v^t - v^\ast}}\epsilon_{\text{est}} + C_2\epsilon^2_{\text{est}},
	\end{eqnarray}
	where $C_0 := \rbr{1 + \frac{2}{1 - 2\lambda^2\frac{\sigma^2_{\min}}{\sigma^2_{\max}}}\rbr{\eta^2\beta^2\rbr{\inner{v^t}{v^t - v^\ast}, y^t}\gamma^2\sigma^2_{\max} - 2\eta\beta\rbr{\inner{v^t}{v^t - v^\ast}, y^t}\gamma\sigma_{\min}\rbr{1 - \lambda}}}$, $C_1 := \eta^2LR^3\sigma^2_{\max} + 2\eta LR^2\sigma_{\max}$, and $C_2 := 2\eta LR^2\sigma_{\max} +  \eta^2L^2R^4\sigma^2_{\max}$.

	Under the ET condition, we are able to pick $\hat x $ and $\hat y$ so that $0 < \gamma\nabla_{\inner{v^t}{\hat x}}\ell\rbr{\inner{v^t}{\hat x}, \hat y} < 2\frac{\rbr{1 - \lambda}\sigma_{\min}}{\eta\sigma^2_{\max}}$, we obtain,
	\begin{eqnarray*}
		C_0  = 1 + \frac{2}{1-2\lambda^2}\rbr{\eta^2\beta^2\rbr{\inner{v^t}{v^t - v^\ast}, y^t}\gamma^2\sigma^2_{\max} - 2\eta\beta\rbr{\inner{v^t}{v^t - v^\ast}, y^t}\gamma\sigma_{\min}\rbr{1 - \lambda}}\le 1.
	\end{eqnarray*}
	With the condition $\forall v\in \Omega_v$, $\nbr{v}\le C_v$ and $\beta\rbr{\inner{v}{x^t}, y^t}\le C_\beta$ holds, as long as we can obtain $\epsilon_{\text{est}} = \Ocal\rbr{\frac{1}{t^2}}$, $\nbr{\xlm{\Gcal^\top (w^{t+1})} - v^\ast}^2$ converges in rate $\Ocal\rbr{\frac{1}{t}}$~\cite{nemirovski2009robust}. In fact, we can achieve better converges rate, \ie, less sample complexity, with more accurate estimation in each iteration. Specifically, we expand the recursion~\eq{eq:recursion},
	\begin{eqnarray*}
		\nbr{\xlm{\Gcal^\top (w^{t+1})} - v^\ast}^2
		&\le& C_0\nbr{\xlm{\Gcal^\top (w^t)} - v^\ast}^2+ \underbrace{C_1\rbr{C_\beta + 2C_v}}_{C_1'}\epsilon_{\text{est}} + C_2\epsilon^2_{\text{est}}\\
		&\le& C_0^2\nbr{\xlm{\Gcal^\top (w)}^{t-1} - v^\ast}^2 + C_0\rbr{C'_1\epsilon_{\text{est}} + C_2\epsilon^2_{\text{est}}}+ C'_1\epsilon_{\text{est}} + C_2\epsilon^2_{\text{est}}\\
		&\le& \cdots\\
		&\le& C_0^{t+1} \nbr{\xlm{\Gcal^\top (w^0)} - v^\ast}^2 + \rbr{\sum_{i=1}^{t} C_0^i}\rbr{C'_1\epsilon_{\text{est}} + C_2\epsilon^2_{\text{est}} }\\
		&=& C_0^{t+1} \nbr{\xlm{\Gcal^\top (w^0)} - v^\ast}^2 + \frac{C_0\rbr{1 - C_0^t}}{1 - C_0}\rbr{C'_1\epsilon_{\text{est}} + C_2\epsilon^2_{\text{est}} }.
	\end{eqnarray*}
	To achieve $\epsilon$-approximation of $v^*$ for student, we may need the number of teaching samples to be
	\begin{equation}\label{eq:teaching_sample}
	T = \rbr{\log\frac{1}{\sqrt{C_0}}}^{-1}\log \frac{2\nbr{\xlm{\Gcal^\top (w^0)} - v^\ast}}{\epsilon}
	\end{equation}
	so that $C_0^{t+1} \nbr{\xlm{\Gcal^\top (w^0)} - v^\ast}^2\le\frac{\epsilon}{2}$, while the number of query samples in each iteration $m$ should satisfy
	
	\begin{equation}\label{eq:testing_sample}
	  \begin{cases}
	    \frac{C_0\rbr{1 - C_0^T}}{1 - C_0}C'_1\epsilon_{\text{est}}\le \frac{C_0}{1 - C_0}C'_1\epsilon_{\text{est}}\le \min\rbr{\frac{\epsilon}{4}, \frac{\lambda\sigma_{\min}}{\sigma_{\max}}\epsilon}	\\
	    \epsilon_{\text{est}}\le \frac{C_1'}{C_2}
  	  \end{cases}
  	  \Rightarrow \epsilon_{\text{est}} \le \min\rbr{\frac{1 - C_0}{C_0C_1'}\min\rbr{\frac{1}{4}, \frac{\lambda\sigma_{\min}}{\sigma_{\max}}}\epsilon, \frac{C_1'}{C_2}}.
	\end{equation}
	Then, the total number of samples will be
	$$
	T\rbr{1 + m\rbr{\epsilon_{\text{est}}, \frac{\delta}{T}}} = \Ocal\rbr{\log\frac{1}{\epsilon}\rbr{1 + m\rbr{{\lambda\epsilon}, \frac{\delta}{\log\frac{1}{\epsilon}}}}}.
	$$
\end{proof}

{\bf Theorem~\ref{thm:sign_F}}
\emph{
	\xlm{Suppose that Assumption~\ref{eq:1bit_assumption} holds. Then} with probability at least $\thickmuskip=2mu \medmuskip=2mu 1 - \delta$, then we can recover $\Gcal^\top(w) \in \RR^d$ with $\tilde\Ocal\rbr{\rbr{d^2 + d\log\frac{1}{\delta}}\log\frac{1}{\epsilon}}$ query samples.
}
\begin{proof}
Similarly, we prove this claim by construction. Basically, we first approximate the $\tilde\alpha = \frac{\xlm{\Gcal^\top (w)}}{\nbr{\xlm{\Gcal^\top (w)}}}$ within $\Omega_\alpha = \cbr{\alpha\in \RR^d, \nbr{\alpha}=1}$, and then, rescale it by $\nbr{\xlm{\Gcal^\top (w)}}$.

In the first stage, we exploit active learning~\cite{balcan2009agnostic}. Obvisouly, $\nbr{v} = 1$, therefore, after $t$-iteration in examination phase, we have
\begin{align*}
\nbr{\alpha_t - \tilde\alpha}^2 = \nbr{\alpha_t}^2 + \nbr{\tilde\alpha}^2 - 2\inner{\alpha_t}{\tilde\alpha} = 2\rbr{1 - \cos\rbr{\alpha_t, \tilde\alpha}} = 2\rbr{1 - \sqrt{1 - \sin^2\rbr{\alpha_t, \tilde\alpha}}},
\end{align*}
therefore,
\begin{align*}
\nbr{\alpha_t - \tilde\alpha}^2\le2\sin\rbr{\alpha_t, \tilde\alpha}.
\end{align*}
which is obtained by applying $\sqrt{1 - x^2}\ge (1 - x)$ when $0\le x\le 1$. Recall $\sin\rbr{\alpha_t, \tilde\alpha} = \Ocal\rbr{\frac{1}{2^t\sqrt{d}}}$, we have
\begin{align*}
\nbr{\alpha_t - \tilde\alpha}^2 = \Ocal\rbr{\frac{1}{2^t\sqrt{d}}},
\end{align*}
which is equivalent that we can approximate $\nbr{\alpha_t - \tilde\alpha}^2\le \epsilon$ with $t = \Ocal\rbr{\log\frac{1}{\epsilon}}$. In each iteration, the active learning make $\tilde\Ocal\rbr{d^2\log d + d\log\frac{1}{\delta}}$ queries, implying the total sample complexity is $\tilde\Ocal\rbr{\rbr{d^2 + d\log\frac{1}{\delta}}\log\frac{1}{\epsilon}}$.

When rescaling, we increase the error by $\nbr{\xlm{\Gcal^\top (w)}}^2$, then, we can set $\epsilon' = \frac{\epsilon}{\nbr{\xlm{\Gcal^\top (w)}}^2}$. When $\nbr{\xlm{\Gcal^\top (w)}}$ is bounded by some constant $C$, which is the case, the sample we needed will be $\tilde\Ocal\rbr{\rbr{d^2 + d\log\frac{1}{\delta}}\log\frac{C^2}{\epsilon}}$ which does not affect the asymptotic sample complexity.
\end{proof}
Plug Theorem~\ref{theorem:recursion_with_error} with Theorem~\ref{thm:sign_F}, we have

{\bf Corollary~\ref{cor:sign_F}}
\emph{\xlm{Suppose that Assumption~\ref{eq:1bit_assumption} holds. Then} then $(\ell,\Gcal)$ is ET with $\Ocal\rbr{\log\frac{1}{\epsilon}}$ teaching samples and $\tilde\Ocal\rbr{\log\frac{1}{\epsilon}\log\frac{1}{\lambda\epsilon}\rbr{d^2 + d\log\frac{\log \frac{1}{\epsilon}}{\delta}}}$ query samples.
}

%%%------------------------------------------------------------------------------------------------
\subsection{Estimation Error Preservation}

{\bf Lemma~\ref{lemma:error_preservation}}
\emph{
	\xlm{Suppose} that $\Gcal$ is a unitary operator\xlm{. If} $\nbr{\Gcal^\top (w^0)  - v^0}\le \epsilon$, then $\nbr{\Gcal^\top (w^{t+1}) - v^{t+1}} \le \epsilon$.
}
\begin{proof}
	This can be checked by induction, assume in $t$-th step, $\nbr{\xlm{\Gcal^\top (w^t)}  - v^t}\le \epsilon$,
	\begin{eqnarray*}
		\nbr{\xlm{\Gcal^\top (w^{t+1})} - v^{t+1}} &=& \nbr{\Gcal (w^t) - \eta\beta_{\inner{v^t}{x_t}}\Gcal^\top\xlm{\Gcal (x)}_t - v^t + \eta\beta_{\inner{v^t}{x_t}}x_t}\\
		& = & \nbr{\xlm{\Gcal^\top (w^t)} - v^t}\le \epsilon.
	\end{eqnarray*}
\end{proof}

%%%------------------------------------------------------------------------------------------------
\subsection{Extension to Combination-based and Pool-based Active Teaching}
%%%------------------------------------------------------------------------------------------------

In this section, we mainly discuss the results for synthesis-based active teaching to combination-based and pool-based active learning.

For {\bf combination-based active teaching}, where both the training samples and query samples are constructed by linear combination of $k$ samples $\Dcal = \cbr{x_i}_{i=1}^k$, we have the following results for exact recovery and approximate recovery in the sense of
\begin{align*}
\xlm{\inner{v_1}{v_2}_{\Dcal}} := \sqrt{v_1^\top \Dcal\rbr{\Dcal^\top\Dcal}^{+}\Dcal^\top v_2},~~\text{and}~~\nbr{v}_{\Dcal}:=\inner{v}{v}_{\Dcal}.
\end{align*}
Note that with the introduced metric, for $v\in \RR^d$, we only consider its component in ${\rm span}\rbr{\Dcal}$ and the components in the null space will be ignored. Therefore, $\forall~ v_1, v_2\in {\rm span}(\Dcal)$ such that $\nbr{v_1}_{\Dcal} = \nbr{v_2}_{\Dcal}$, we have $\thickmuskip=2mu \medmuskip=2mu  v_1^\top x = v_2^\top x = \xlm{\inner{v_1}{x}_{\Dcal}}$ for all $x\in\RR^d$. For notational convenience, we omit the subscript $\Dcal$ for the analysis in this section.

{\bf Corollary~\ref{cor:combination_exact}}
\emph{
Suppose the student answers questions in query phase via $F(\cdot) = I(\cdot)$ or $F(\cdot) = S(\cdot)$ and $\Gcal^\top (w^0), v^\ast\in {\rm span}\rbr{\Dcal}$. Then $(\ell,\Gcal)$ is ET with $\Ocal\rbr{\log\frac{1}{\epsilon}}$ teaching samples and ${\rm rank}(\Dcal)$ query samples via exact recovery.
}

{\bf Corollary~\ref{cor:combination_approximate}}
\emph{
Suppose Assumption~\ref{eq:1bit_assumption} holds, the student answers questions in query phase via $F(\cdot) = I(\cdot)$ or $F(\cdot) = S(\cdot)$ and $\Gcal^\top (w^0), v^\ast\in {\rm span}\rbr{\Dcal}$. Then $(\ell,\Gcal)$ is ET with $\Ocal\rbr{\log\frac{1}{\epsilon}}$ teaching samples and $\tilde\Ocal\rbr{\log\frac{1}{\epsilon}\log\frac{1}{\lambda\epsilon}\rbr{d^2 + d\log\frac{\log \frac{1}{\epsilon}}{\delta}}}$ query samples via approximate recovery.
}

The proof for these two corollaries are straightforward since under the condition that $\xlm{\Gcal^\top (w^0)}, v^\ast\in {\rm span}\rbr{\Dcal}$, every teaching sample will be in ${\rm span}\rbr{\Dcal}$, so that the $\xlm{\Gcal^\top (w^t)}$ and $v^t$ are also in ${\rm span}\rbr{\Dcal}$. Therefore, we can reduce such setting to synthesis-based active teaching with essential dimension as ${\rm rank}(\Dcal)$. Then, the conclusions are achieved.

For {\bf rescaled pool-based active teaching}, where the teacher can only pick samples from a prefixed sample candidates pool, $\Dcal = \cbr{x_i}_{i=1}^k$, for teaching and query. We will still evaluate using the same metric $\nbr{\cdot}_{\Dcal}$ defined above (omit subscript $\Dcal$ for convenience). We first discuss the exact recovery case.

{\bf Theorem~\ref{thm:pool_exact}}
\emph{
\xlm{Suppose} the student answers questions in the exam phase via $\thickmuskip=2mu \medmuskip=2mu F(\cdot) = I(\cdot)$ or $\thickmuskip=2mu \medmuskip=2mu F(\cdot) = S(\cdot)$ and $\thickmuskip=2mu \medmuskip=2mu \Gcal^\top (w^0), v^\ast\in {\rm span}\rbr{\Dcal}$. If $\forall~ \Gcal^\top (w) \in {\rm span}(D)$, there \xlm{exist} $(x, y)\in \Dcal\times\Ycal$ and $\gamma$ such that for $\thickmuskip=2mu \medmuskip=2mu \hat x = \frac{\gamma\nbr{\Gcal^\top (w) - v^\ast}_{\Dcal}}{\nbr{x}_{\Dcal}}x, \, \hat y = y$, we have
\begin{align*}
0\le \gamma \nabla_{\inner{v^t}{\hat x}}\ell\rbr{\inner{v^t}{\hat x}, \hat y}\le \frac{2\Vcal\rbr{\Xcal}\sigma_{\min}}{\eta\sigma^2_{\max}},
\end{align*}
then $(\ell,\Gcal)$ is ET with $\Ocal\rbr{\log\frac{1}{\epsilon}}$ teaching samples and ${\rm rank}(\Dcal)$ query samples.
}
\begin{proof}
Under the conditions that $\xlm{\Gcal^\top (w^0)}, v^\ast\in {\rm span}\rbr{\Dcal}$, with the same argument, in each iteration, both $\xlm{\Gcal^\top (w^t)}$ and $v^t$ are in ${\rm span}\rbr{\Dcal}$. Therefore, as long as we pick ${\rm rank}(\Dcal)$ independent samples from $\Dcal$ as query samples, we can recover any $v\in {\rm span}\rbr{\Dcal}$ in the sense of the introduced metric.
For the training sample, due to the restriction in selecting samples, we need to recheck the recursion. We follow the proof for rescaled pool-based omniscient teaching in~\cite{liu2017iterative}. Specifically, at $t$-step, as the loss is exponentially synthesis-based teachable with $\gamma$, therefore, we have the virtually constructed sample $\cbr{x_v, y_v}$ where $x_v = \gamma \rbr{\xlm{\Gcal^\top (w^t)} - v^\ast}$ with $\gamma$ satisfying the condition of exponentially synthesis-based active teachability, we first rescale the candidate pool $\Xcal$ such that
$$
\forall x\in \Xcal, \gamma_x\nbr{x} =\nbr{x_v} = \gamma\nbr{\xlm{\Gcal^\top (w^t)} - v^\ast}.
$$
We denote the rescaled candidate pool as $\Xcal_t$, under the condition of rescalable pool-based teachability, there is a sample $\cbr{\hat x, \hat y}\in \Xcal \times \Ycal$ with scale factor $\hat \gamma$ such that
\begin{align*}
&\min_{\rbr{x, y}\in \Xcal_t\times \Ycal} \eta^2 \|\Gcal^\top \nabla_{w^t} \ell\rbr{\inner{w^t}{\hat\gamma\Gcal(x)}, y}\|^2 - 2\eta\inner{\xlm{\Gcal^\top (w^t)} - v^\ast}{\Gcal^\top\nabla_{w^t} \ell\rbr{\inner{w^t}{\hat\gamma\Gcal(x)}, y}}\\
&\le \eta^2 \nbr{\beta{\rbr{\inner{w^t}{\hat\gamma\Gcal(\hat x)}, \hat y}}\Gcal^\top\Gcal(\hat\gamma\hat x)}^2 - 2\eta\beta\rbr{\inner{w^t}{\hat\gamma\Gcal(\hat x)}, \hat y}\inner{\hat \gamma \Gcal^\top\Gcal \hat x}{\xlm{\Gcal^\top (w^t)} - v^\ast}.
\end{align*}
We decompose the $\hat\gamma \hat x = a x_v + {x_v}_\perp$ with $a = \frac{\inner{\hat\gamma \hat x}{{x_v}}}{\nbr{{x_v}}^2}$. and ${x_v}_\perp = \hat\gamma \hat x - a{x_v}$. Then, we have
\begin{align*}
&\min_{\rbr{x, y}\in \Xcal_t\times \Ycal} \eta^2 \|\Gcal^\top \nabla_{w^t} \ell\rbr{\inner{w^t}{\Gcal(x)}, y}\|^2 - 2\eta\inner{\xlm{\Gcal^\top (w^t)} - v^\ast}{\Gcal^\top\nabla_{w^t} \ell\rbr{\inner{w^t}{\Gcal(x)}, y}}\\
&\le \eta^2 \beta^2{\rbr{\inner{w^t}{\hat\gamma\Gcal(\hat x)}, \hat y}}\nbr{\hat\gamma\Gcal^\top\Gcal(\hat x)}^2 - 2\eta\beta\rbr{\inner{w^t}{\hat\gamma\Gcal(\hat x)}, \hat y}\inner{\hat \gamma \Gcal^\top\Gcal \hat x}{\xlm{\Gcal^\top (w^t)} - v^\ast}\\
&\le\eta^2 \beta^2{\rbr{\inner{w^t}{\hat\gamma\Gcal(\hat x)}, \hat y}}\gamma^2\sigma^2_{\max}\nbr{\xlm{\Gcal^\top (w^t)} - v^\ast}^2  - 2\eta\beta\rbr{\inner{w^t}{\hat\gamma\Gcal(\hat x)}, \hat y}\sigma_{\min}\inner{a {x_v} + {x_v}_\perp}{\xlm{\Gcal^\top (w^t)} - v^\ast} \\
&= \eta^2 \beta^2{\rbr{\inner{w^t}{\hat\gamma\Gcal(\hat x)}, \hat y}}\gamma^2\sigma^2_{\max}\nbr{\xlm{\Gcal^\top (w^t)} - v^\ast}^2  - 2\eta\beta\rbr{\inner{w^t}{\hat\gamma\Gcal(\hat x)}, \hat y}\sigma_{\min}a\nbr{\xlm{\Gcal^\top (w^t)} - v^\ast}^2 \\
\end{align*}
Under the condition
\begin{align*}
0\le \gamma \beta{\rbr{\inner{w^t}{\gamma\frac{\nbr{\xlm{\Gcal^\top (w^t)} - v^\ast}}{\nbr{x}}\Gcal(x)}, y}}\le \frac{2\Vcal\rbr{\Xcal}\sigma_{\min}}{\eta\sigma^2_{\max}},
\end{align*}
we have the recursion
\begin{align*}
\nbr{\xlm{\Gcal^\top (w^{t+1})} - v^\ast}^2\le r\rbr{\eta, \gamma, \Gcal, \Vcal\rbr{\Xcal}} \nbr{\xlm{\Gcal^\top (w^{t})} - v^\ast}^2,
\end{align*}
where $r\rbr{\eta, \gamma, \Gcal, \Vcal\rbr{\Xcal}}= {\max\cbr{1 + \eta^2\mu\rbr{\gamma}^2\sigma^2_{\max} - 2\eta\mu\rbr{\gamma}\sigma_{\min} \Vcal(\Xcal), 1 + \eta^2\nu\rbr{\gamma}^2\sigma^2_{\max} - 2\eta\nu\rbr{\gamma}\sigma_{\min} \Vcal(\Xcal)}}$ and $0\le r\rbr{\eta, \gamma, \Gcal, \Vcal\rbr{\Xcal}} <1$, with $\nu\rbr{\gamma} = \min_{w, \hat x\in\Xcal, \hat y\in \Ycal} \gamma\beta{\rbr{\inner{w^t}{\gamma\frac{\nbr{\xlm{\Gcal^\top (w^t)} - v^\ast}}{\nbr{x}}\Gcal(x)}, y}}>0$ and $\mu\rbr{\gamma} = \max_{w, \hat x\in \Xcal, \hat y\in \Ycal}\gamma\beta{\rbr{\inner{w^t}{\gamma\frac{\nbr{\xlm{\Gcal^\top (w^t)} - v^\ast}}{\nbr{x}}\Gcal(x)}, y}}<\frac{2\Vcal\rbr{\Xcal}\sigma_{\min}}{\eta\sigma^2_{\max}}$. Therefore, the algorithm converges exponentially
$$
\nbr{\xlm{\Gcal^\top (w^{t})} - v^\ast}_2 \le {r\rbr{\eta, \gamma, \Gcal, \Vcal\rbr{\Xcal}}}^{t/2}\nbr{\xlm{\Gcal^\top (w^{t})} - v^\ast}_2.
$$
In sum, the student needs $2\rbr{\log\frac{1}{r\rbr{\eta, \gamma, \Gcal, \Vcal(\Xcal)}}}^{-1}\log\frac{\|\xlm{\Gcal^\top (w^0)} - v^\ast\|}{\epsilon}$ teaching samples and ${\rm rank}(\Dcal)$ query samples to achieve an $\epsilon$-approximation of $v^\ast$.
\end{proof}

For approximate recovery case, the active learning is no longer able to achieve the required accuracy for estimating of the student parameters with the restricted sample pool. Therefore, the algorithm may not achieve exponential teaching. We will leave this as an open problem.

\end{appendix}

\clearpage
\section{Experimental Details}\label{appendix:exp}
For synthetic data, we generate training data $(\bm{x}_i,y)$ where each entry in $\bm{x}_i$ is Gaussian distributed and $y = \inner{\bm{w}^*}{\bm{x}_i}+\epsilon$ where $\epsilon$ is a Gaussian distributed noise for the LSR learner. For the LR learner, $\{\mathcal{X}_1,+1\}$ and $\{\mathcal{X}_2,-1\}$ where $\bm{x}_i\in\mathcal{X}_1$ is Gaussian distributed in each entry and $+1,-1$ are the labels. Specifically, we use the 50-dimension data that is Gaussian distributed with $(0.5, \cdots,0.5)$ (label +1) and $(-0.5,\cdots,-0.5)$ (label -1) as the mean and identity matrix as the covariance matrix. We generate 1000 training data points for each class. Learning rate for the same feature space is 0.0001, $\lambda$ for regularization term is set as 0.00005. For the operator $\Gcal$ that maps between teacher's and student's spaces, we mostly use an orthogonal transformation in experiments. In MNIST dataset, we use full training set of digits 7 and 9 and extract 24-dim projective random features from the raw $32\times 32$ images. We use the full testing set to evaluate the 7/9 classification accuracy.

\section{More Experiments: LR Learner with $\bm{F(z)=S(z)}$}\label{appendix:more}
\begin{figure*}[h]
	\centering
	\footnotesize
	\renewcommand{\captionlabelfont}{\footnotesize}
	\includegraphics[width=0.99\linewidth]{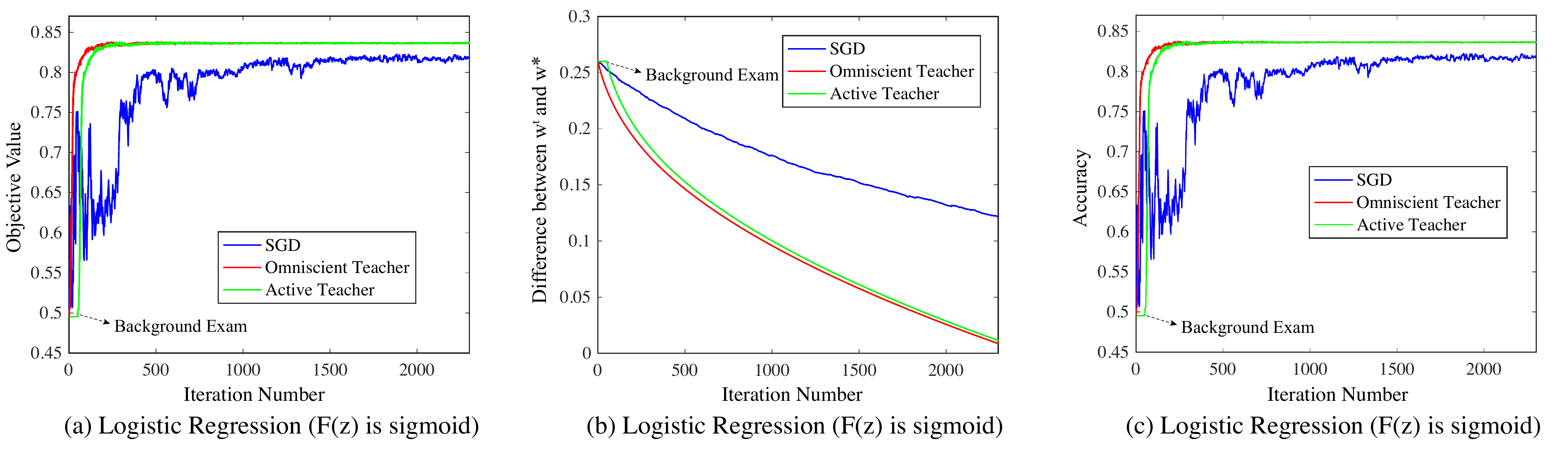}
	\vspace{-1mm}
	\caption{\footnotesize The convergence performance of random teacher (SGD), omniscient teacher and active teacher in MNIST 7/9 classification. We evaluate the LR learner with $F(z)=S(z)$ here.}\label{mnist_f121}
	%\vspace{-1.5mm}
\end{figure*}

For the LR learner that uses the sigmoid function as feedbacks, one could clearly see that the experimental results match our theoretical analysis in case of the exact recovery of the ideal virtual learner. The active teacher is able to achieve the same performance as the omniscient teacher after the ``background exam'', and converges much faster than the SGD. In fact, the active teacher and the omniscient teacher should achieve the same convergence speed without consideration of numerical errors. Moreover, the empirical results indicate that the teacher tends to pick easy examples first and difficult examples later. In iterative machine teaching, the difficulty level of an example is essentially the distance between the example and the decision boundary. Interestingly, deeply learned features also exhibit similar difficulty level in terms of the norm of the feature~\cite{liu2018decoupled,liu2017deep}, which may be useful for improving the convergence of deep models (\eg, the norm fo deeply learned features can be used as a form of difficulty indicator in curriculum learning and iterative machine teaching).

\newpage
\section{Analysis and Experiments of the Learner with Forgetting Behavior}\label{forgetting}
\subsection{Modeling the forgetting behavior}
We model the forgetting behavior of the learner by adding a deviation to the learned parameter in each iteration of updating the learner. Specifically in each iteration, the learner will update its model in its feature space with 
\begin{equation}
w^{t+1}=w^t+\nabla_w\ell(\langle w^t,x\rangle,y) + \epsilon_t
\end{equation}
where $\epsilon_t$ is a random deviation vector. The larger the deviation is, the more the learner forgets. $\epsilon_t$ can be modeled in a time-variant fashion, or simply using a fixed probability distribution. There will be a number of ways to model the deviation. For simplicity, we only consider a Gaussian distribution with zero mean and fixed variance here. Throughout this section, we mainly study the case where the teacher and learner share the same feature space when the learner has the forgetting behavior. It could help us simplify the problem, but it also more clearly shows the superiority of the active teacher because the setting is comparable to the omniscient teacher.
\subsection{The exponential teachability of the learner with forgetting behavior}
\begin{figure*}[h]
	\centering
	\footnotesize
	\renewcommand{\captionlabelfont}{\footnotesize}
	\includegraphics[width=0.85\linewidth]{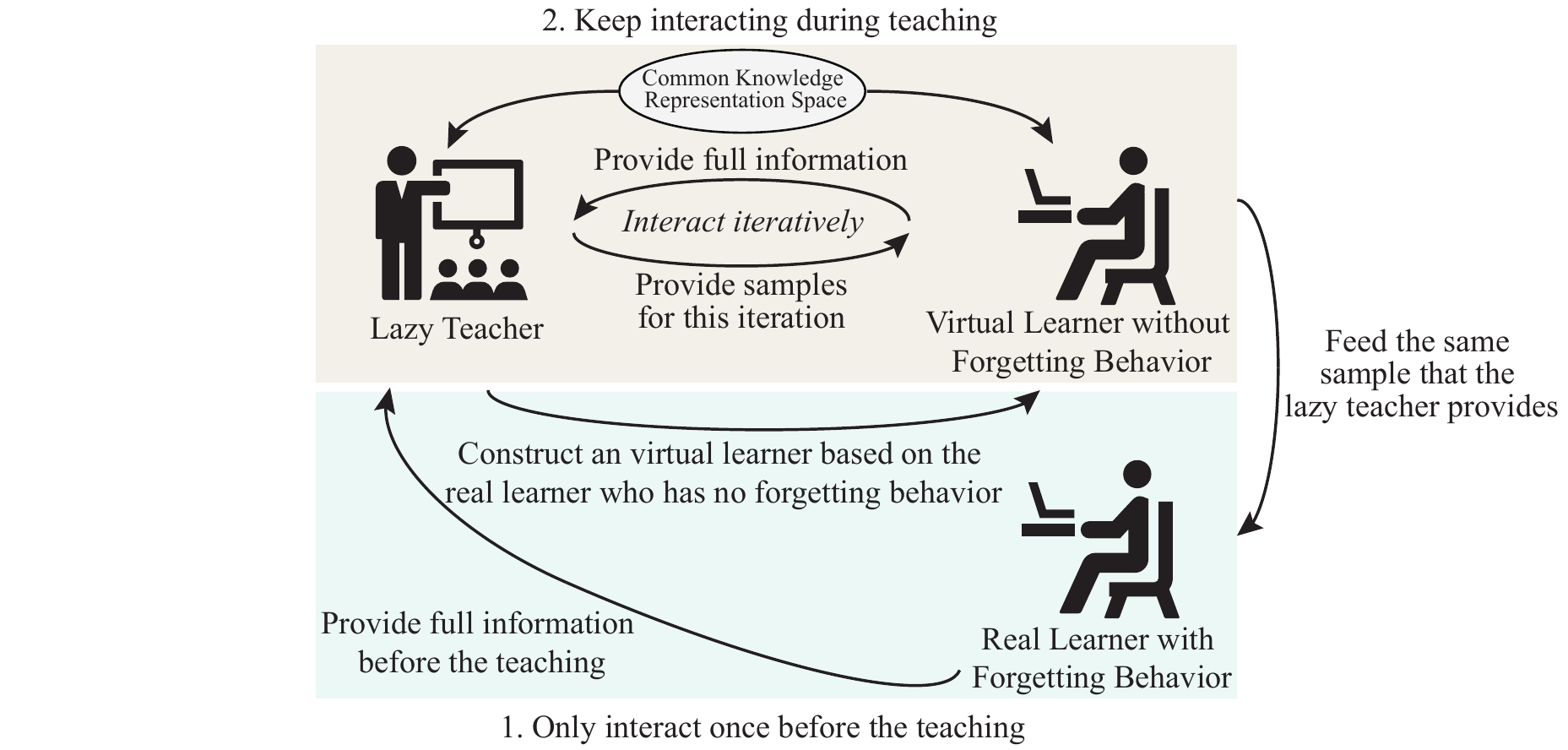}
	\vspace{-1mm}
	\caption{\footnotesize An illustrative overview of the lazy teacher.}\label{lazy}
	%\vspace{-1.5mm}
\end{figure*}

Before delving deep into the exponential teachability of the learner with forgetting behavior, we first define a lazy teacher model. The lazy teacher model works essentially similar to the omniscient teacher, except that the lazy teacher will first construct a virtual learner before the teaching and will not observe the status of the learner during iteration. Specifically, the lazy teacher will first construct a virtual learner without forgetting behavior based on the initial status (information) from the real learner. Then the lazy teacher will pick samples based on the observation from the virtual learner and will feed the same samples to the real learner. One can notice that if the real learner has no forgetting behavior, the lazy teacher will be identical to the omniscient teacher. An overview of the lazy teacher is given in Fig.~\ref{lazy}.

For the learner guided by the active teacher to achieve ET, it requires the sample complexity of the active learning to be $\Ocal(\log \frac{1}{\epsilon})$, as shown in Theorem \ref{theorem:recursion_with_error}. It is obvious that the deviation error $\epsilon_t$ of a forgetting learner can not converge to a small enough value. Therefore, the forgetting learner can not achieve ET with the lazy teacher, because the the deviation error can not be controlled by the lazy teacher. In contrast, the forgetting learner can still achieve ET with our proposed active teacher, because the deviation error can also be estimated by the active query strategy. In other words, the active teacher is still able to estimate accurate enough current parameters of the forgetting learner, which also prevents the deviation error to propagate over iterations.

\subsection{Experiments}
\begin{figure*}[h]
	\centering
	\footnotesize
	\renewcommand{\captionlabelfont}{\footnotesize}
	\includegraphics[width=0.7\linewidth]{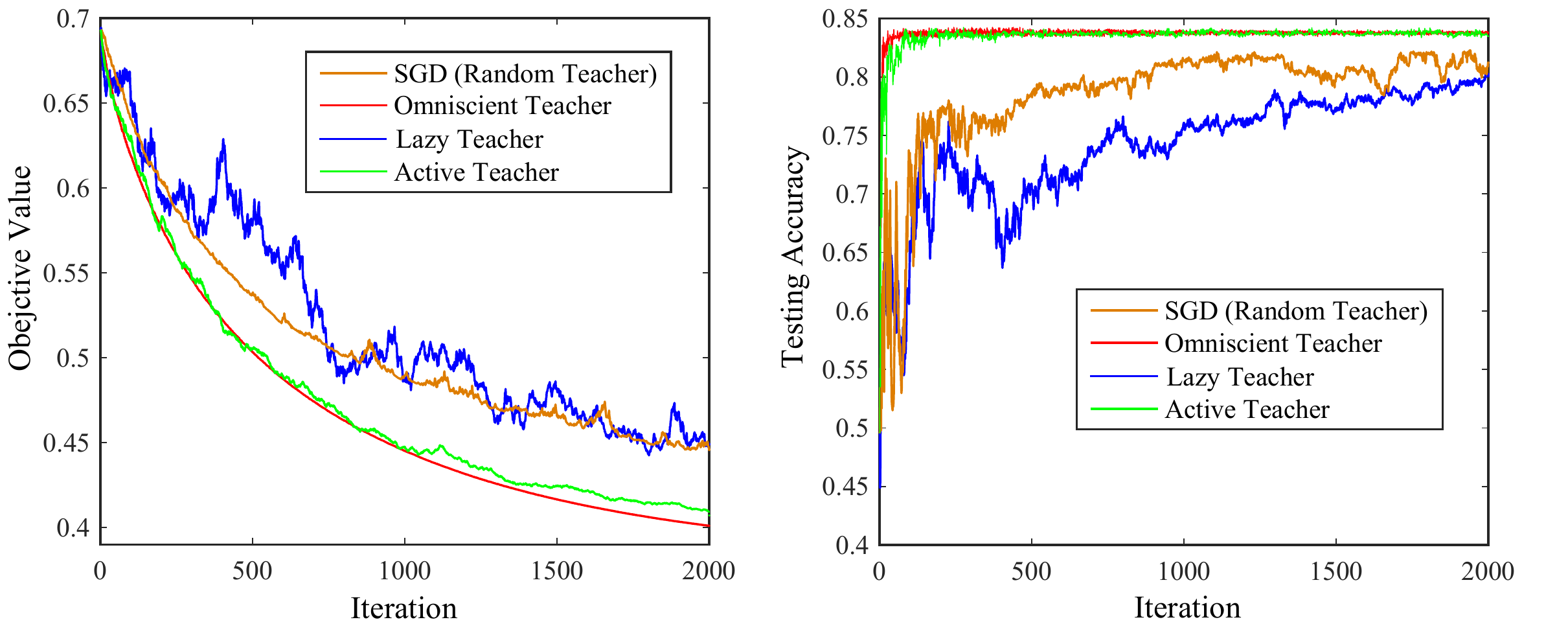}
	\vspace{-1mm}
	\caption{\footnotesize The convergence performance of random teacher (SGD), omniscient teacher, lazy teacher and active teacher in MNIST 7/9 binary classification.}\label{forget_exp}
	%\vspace{-1.5mm}
\end{figure*}

We perform an experiment on MNIST dataset to show how the forgetting behavior will affect the fast convergence, and also compare our active teacher with the lazy teacher. We still use the binary classification for digit 7 and 9 for our experiment. The experimental setting for the MNIST dataset is similar to Section~\ref{mnist_exp_sect} except that we only use one random projection to extract the features, which means that the teacher and the learner share the same feature space. We could see from Fig.~\ref{forget_exp} that the forgetting behavior will greatly affect the convergence of the lazy learner, while the lazy learner have the same convergence speedup with the omniscient teacher if the learner has no forgetting behavior. Most importantly, our active teacher can well address this forgetting behavior and provide significant convergence speedup. This experiment also partially validates that it is very reasonable in real-world education to make students take exam. If the teacher model can not well estimate or have access to the current parameter of the student model, then the entire teaching will very possibly fail (\ie, similar to or even worse than the random teacher).
\par
\textbf{Experimental settings.} We perform the experiment on MNIST dataset with digit 7/9 binary classification. The 24-dim feature is computed by random projection from raw pixels. The learner will provide $F(z)=\textnormal{sign}(z)$ as feedbacks. For fairness, the learning rates for all method are the same. 

\end{document}